\documentclass[10pt,twocolumn,letterpaper]{article}

\usepackage[pagenumbers]{cvpr} 

\usepackage{graphicx}
\usepackage{amsmath}
\usepackage{amssymb}
\usepackage{booktabs}
\usepackage{multirow}

\newcommand{\tight}[1]{\hspace{-.7mm}#1\hspace{-.7mm}}

\usepackage{color}
\usepackage[pagebackref=true,breaklinks=true,letterpaper=true,colorlinks,citecolor=citecolor,bookmarks=false]{hyperref}
\definecolor{citecolor}{RGB}{30,130,255}
\definecolor{palegreen}{RGB}{210,240,200}

\usepackage{soul}

\usepackage[capitalize]{cleveref}
\crefname{section}{Sec.}{Secs.}
\Crefname{section}{Section}{Sections}
\Crefname{table}{Table}{Tables}
\crefname{table}{Tab.}{Tabs.}

\makeatletter
\def\blfootnote{\gdef\@thefnmark{}\@footnotetext}
\makeatother

\begin{document}

 \title{InstructPix2Pix: Learning to Follow Image Editing Instructions\vspace{-3.5mm}}

\author{%
\hspace{4mm}Tim Brooks*%
\hspace{8mm}Aleksander Holynski*%
\hspace{7mm} Alexei A. Efros%
\vspace{1.7mm}\\University of California, Berkeley%
}

\twocolumn[{%
\renewcommand\twocolumn[1][]{#1}%
\maketitle

\vspace{-7.5mm}
\begin{center}
    \includegraphics[width=\linewidth]{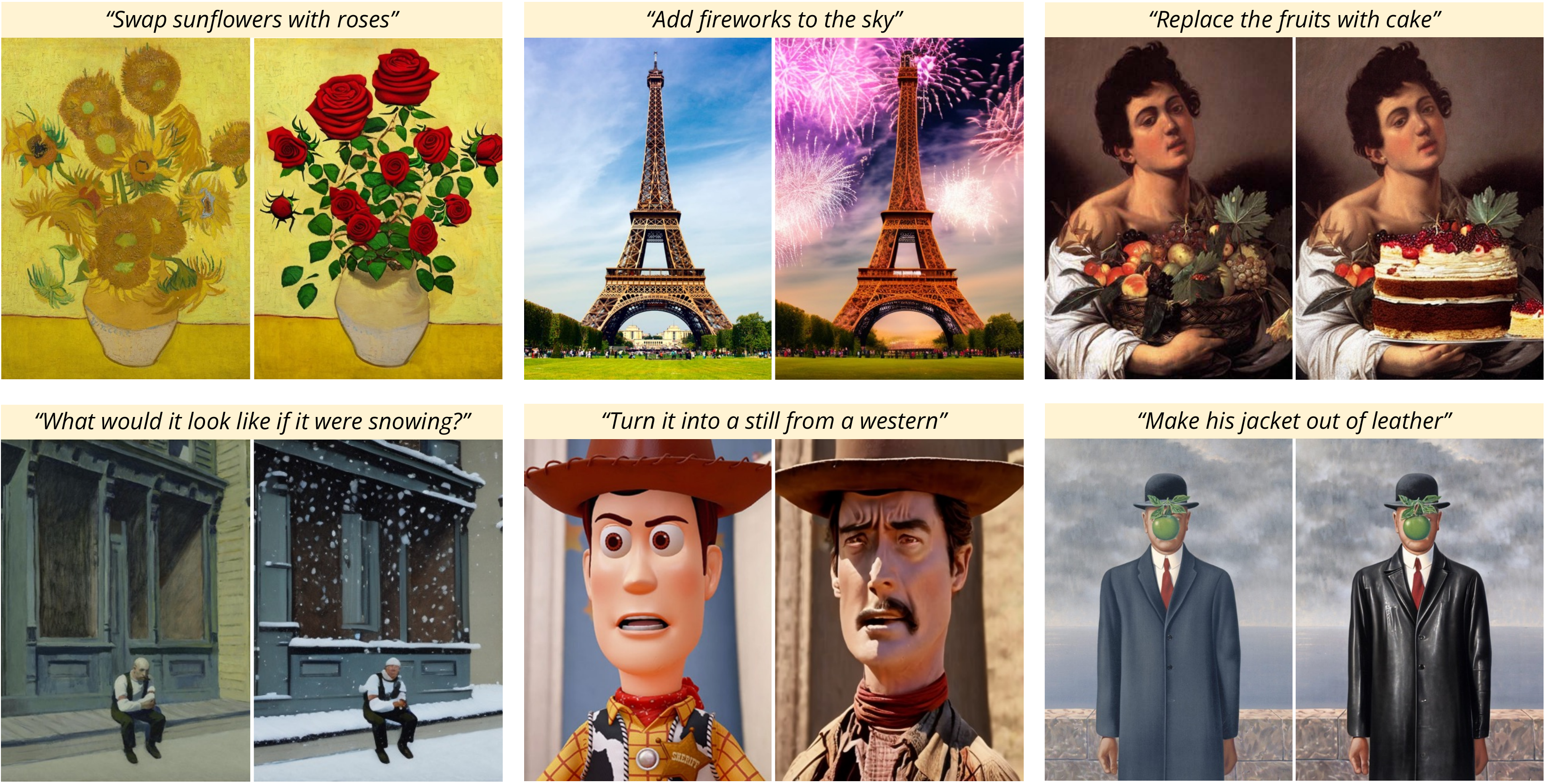}%
    \vspace{-2.3mm}
    \captionof{figure}{
    Given {\bf an image} and {\bf an instruction} for how to edit that image, our model performs the appropriate edit. Our model does not require full descriptions for the input or output image, and edits images in the forward pass without per-example inversion or fine-tuning.}
    \label{fig:teaser}
    \vspace*{2.7mm}
\end{center}%
}]

\begin{abstract}

\vspace*{-2.9mm}
We propose a method for editing images from human instructions: given an input image and a written instruction that tells the model what to do, our model follows these instructions to edit the image.
To obtain training data for this problem, we combine the knowledge of two large pretrained models---a language model (GPT-3) and a text-to-image model (Stable Diffusion)---to generate a large dataset of image editing examples.
Our conditional diffusion model, InstructPix2Pix, is trained on our generated data, and generalizes to real images and user-written instructions at inference time. Since it performs edits in the forward pass and does not require per-example fine-tuning or inversion, our model edits images quickly, in a matter of seconds. 
We show compelling editing results for a diverse collection of input images and written instructions. 
\vspace*{-4.8mm}

\blfootnote{\hspace{-2em}*Denotes equal contribution\\\hspace{-2em}More results on our project page: \href{http://timothybrooks.com/instruct-pix2pix}{timothybrooks.com/instruct-pix2pix}}

\end{abstract}

\section{Introduction}
\label{sec:intro}
\vspace{-0.5mm}

We present a method for teaching a generative model to follow human-written instructions for image editing. Since training data for this task is difficult to acquire at scale, we propose an approach for generating a paired dataset that combines multiple large models pretrained on different modalities: a large language model (GPT-3~\cite{brown2020language}) and a text-to-image model (Stable Diffusion~\cite{rombach2022high}). 
These two models capture complementary knowledge about language and images that can be combined to create paired training data for a task spanning both modalities.

Using our generated paired data, we train a conditional diffusion model that, given an input image and a text instruction for how to edit it, generates the edited image. 
Our model directly performs the image edit in the forward pass, and does not require any additional example images, full descriptions of the input/output images, or per-example finetuning. 
Despite being trained entirely on synthetic examples (i.e., both generated written instructions and generated imagery), our model achieves zero-shot generalization to both arbitrary \emph{real} images and natural human-written instructions. 
Our model enables intuitive image editing that can follow human instructions to perform a diverse collection of edits: replacing objects, changing the style of an image, changing the setting, the artistic medium, among others. Selected examples can be found in Figure~\ref{fig:teaser}. 

\begin{figure*}[t]
    \centering
    \includegraphics[width=\linewidth]{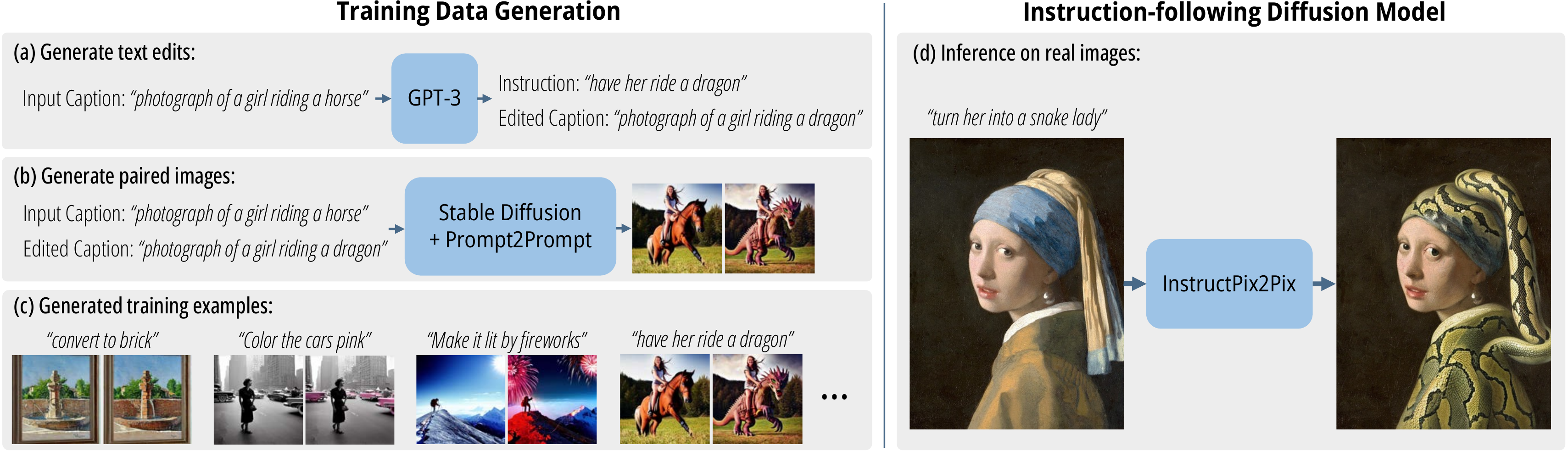}
    \vspace{-.2in}
    \caption{Our method consists of two parts: generating an image editing dataset, and training a diffusion model on that dataset. (a) We first use a finetuned GPT-3 to generate instructions and edited captions. (b) We then use StableDiffusion~\cite{rombach2022high} in combination with Prompt-to-Prompt~\cite{hertz2022prompt} to generate pairs of images from pairs of captions. We use this procedure to create a dataset (c) of over 450,000 training examples. (d) Finally, our InstructPix2Pix diffusion model is trained on our generated data to edit images from instructions. At inference time, our model generalizes to edit real images from human-written instructions.}
    \label{fig:method}
\end{figure*}

\section{Prior work}
\label{sec:prior_work}

\paragraph{Composing large pretrained models}
Recent work has shown that large pretrained models can be combined to solve multimodal tasks that no one model can perform alone, such as image captioning and visual question answering (tasks that require the knowledge of both a large language model and a text-image model). Techniques for combining pretrained models include joint finetuning on a new task~\cite{alayrac2022flamingo,mokady2021clipcap,wang2021simvlm,li2019visualbert}, communication through prompting~\cite{zeng2022socratic,tiong2022plug}, composing probability distributions of energy-based models~\cite{du2020compositional,liu2022compositional},  guiding one model with feedback from another~\cite{tewel2021zero}, and iterative optimization~\cite{li2022composing}. Our method is similar to prior work in that it leverages the complementary abilities of two pretrained models---GPT\nobreakdash-3~\cite{brown2020language}) and Stable Diffusion~\cite{rombach2022high}---but differs in that we use these models to generate paired multi-modal training data.

\vspace{-2.2mm}
\paragraph{Diffusion-based generative models}
Recent advances in diffusion models~\cite{pmlrv37sohldickstein15} have enabled state-of-the-art image synthesis~\cite{song2019generative,ho2020denoising,dhariwal2021diffusion,ho2022cascaded,saharia2022image,saharia2022palette} as well as generative models of other modalities such as video~\cite{ho2022video,singer2022make}, audio~\cite{kong2020diffwave}, text~\cite{li2022diffusion} and network parameters~\cite{peebles2022learning}. Recent text-to-image diffusion models~\cite{nichol2021glide,rombach2022high,ramesh2022hierarchical,saharia2022photorealistic} have shown to generate realistic images from arbitrary text captions.

\vspace{-2.5mm}
\paragraph{Generative models for image editing}
Image editing models traditionally targeted a single editing task such as style transfer~\cite{gatys2015neural,gatys2016image} or translation between image domains~\cite{pix2pix2017,CycleGAN2017,huang2018munit,liu2019few,ojha2021few-shot-gan}. Numerous editing approaches invert~\cite{abdal2019image2stylegan,abdal2020image2stylegan++,alaluf2022hyperstyle,epstein2022blobgan} or encode~\cite{tov2021designing,richardson2021encoding,chai2021latent} images into a latent space (e.g., StyleGAN~\cite{karras2019style,karras2020analyzing}) where they can be edited by manipulating latent vectors. Recent models have leveraged CLIP~\cite{radford2021learning} embeddings to guide image editing using text~\cite{kwon2022clipstyler,Patashnik_2021_ICCV,zheng2022bridging,gal2022stylegan,nichol2021glide,kim2022diffusionclip,avrahami2022blended, crowson2022vqgan}. We compare with one of these methods, Text2Live~\cite{bar2022text2live}, an editing method that optimizes for an additive image layer that maximizes a CLIP similarity objective. 

Recent works have used pretrained text-to-image diffusion models for image editing~\cite{meng2021sdedit,kawar2022imagic,avrahami2022blended,ramesh2022hierarchical,hertz2022prompt}. While some text-to-image models natively have the ability to edit images (e.g., DALLE-2 can create variations of images, inpaint regions, and manipulate the CLIP embedding~\cite{ramesh2022hierarchical}), using these models for \emph{targeted} editing is non-trivial, because in most cases they offer no guarantees that similar text prompts will yield similar images. Recent work by Hertz \textit{et al.}~\cite{hertz2022prompt} tackles this issue with Prompt-to-Prompt, a method for assimilating the generated images for similar text prompts, such that isolated edits can be made to a generated image. We use this method in generating training data. To edit non-generated (i.e., real) imagery, SDEdit~\cite{meng2021sdedit} uses a pretrained model to noise and denoise an input image with a new target prompt. We compare with SDEdit as a baseline.
Other recent works perform local inpainting given a caption and user-drawn mask~\cite{ramesh2022hierarchical,avrahami2022blended}, generate new images of a specific object or concept learned from a small collection of images~\cite{gal2022image,ruiz2022dreambooth}, or perform editing by inverting (and fine-tuning) a single image, and subsequently regenerating with a new text description~\cite{kawar2022imagic}. In contrast to these approaches, our model takes only a single image and an instruction for how to edit that image (i.e., not a full description of any image), and performs the edit directly in the forward pass without need for a user-drawn mask, additional images, or per-example inversion or finetuning.

\vspace{-2.7mm}
\paragraph{Learning to follow instructions}
Our method differs from existing text-based image editing works~\cite{meng2021sdedit,ruiz2022dreambooth, gal2022image,hertz2022prompt,kawar2022imagic,bar2022text2live} in that it enables editing from {\em instructions} that tell the model what action to perform, as opposed to text labels, captions or descriptions of input/output images.
A key benefit of following editing instructions is that the user can just tell the model exactly what to do in natural written text. There is no need for the user to provide extra information, such as example images or descriptions of visual content that remains constant between the input and output images. Instructions are expressive, precise, and intuitive to write, allowing the user to easily isolate specific objects or visual attributes to change.
Our goal to follow written image editing instructions is inspired by recent work teaching large language models to better follow human instructions for language tasks~\cite{ouyang2022training,mishra2021cross,wei2021finetuned}.

\vspace{-2.7mm} 
\paragraph{Training data generation with generative models}
Deep models typically require large amounts of training data. Internet data collections are often suitable, but may not exist in the form necessary for supervision, e.g., paired data of particular modalities.  As generative models continue to improve, there is growing interest in their use as a source of cheap and plentiful training data for downstream tasks~\cite{shrivastava2017learning,ravuri2019classification,viazovetskyi2020stylegan2,tritrong2021repurposing,li2022bigdatasetgan,peebles2022gan}. In this paper, we use two different off-the-shelf generative models (language, text-to-image) to produce training data for our editing model.

    
\section{Method}
\label{sec:method}

We treat instruction-based image editing as a supervised learning problem: (1) first, we generate a paired training dataset of text editing instructions and images before/after the edit (Sec.~\ref{subsec:data}, Fig.~\ref{fig:method}a-c), then (2) we train an image editing diffusion model on this generated dataset (Sec.~\ref{subsec:model}, Fig~\ref{fig:method}d). Despite being trained with generated images and editing instructions, our model is able to generalize to editing \emph{real} images using arbitrary human-written instructions. See Fig.~\ref{fig:method} for an overview of our method.

\subsection{Generating a Multi-modal Training Dataset}
\label{subsec:data}

We combine the abilities of two large-scale pretrained models that operate on different modalities---a large language model~\cite{brown2020language} and a text-to-image model~\cite{rombach2022high}---to generate a multi-modal training dataset containing text editing instructions and the corresponding images before and after the edit. In the following two sections, we describe in detail the two steps of this process. In Section~\ref{subsubsec:gpt}, we describe the process of fine-tuning GPT-3~\cite{brown2020language} to generate a collection of text edits: given a prompt describing an image, produce a text instruction describing a change to be made and a prompt describing the image after that change (Figure~\ref{fig:method}a). Then, in Section~\ref{subsubsec:text2img}, we describe the process of converting the two text prompts (i.e., before and after the edit) into a pair of corresponding images using a text-to-image model~\cite{rombach2022high} (Figure~\ref{fig:method}b).

\vspace{-1mm} 
\subsubsection{Generating Instructions and Paired Captions}
\label{subsubsec:gpt}

\sethlcolor{palegreen}
\begin{table*}
  \centering
  \resizebox{\linewidth}{!}{%
  \begin{tabular}{@{}rp{5cm}lp{5.5cm}@{}}
    \toprule
    & \bf{Input LAION caption} &
     \bf{Edit instruction} &
     \bf{Edited caption} \\
    
    \midrule
    \multirow{4}{2.7cm}{\textbf{Human-written\\(700 edits)}} & \small  \textit{Yefim Volkov, Misty Morning} &
    \small \textit{make it afternoon} &
     \small  \textit{Yefim Volkov, Misty Afternoon} \\ \cline{2-4}
    & \small  \textit{girl with horse at sunset} &
     \small  \textit{change the background to a city} &
     \small  \textit{girl with horse at sunset in front of city} \\ \cline{2-4}
    & \small  \textit{painting-of-forest-and-pond} &
     \small  \textit{Without the water.} &
     \small  \textit{painting-of-forest} \\ \cline{2-4}
    & \small  {...} &
     \small  {...} &
     \small  {...} \\
    \bottomrule
    \toprule
    \multirow{7}{2.7cm}{\textbf{GPT-3 generated\\($>$450,000 edits)}} & \small \textit{Alex Hill, Original oil painting on canvas, Moonlight Bay} &
     \small \hl{\textit{in the style of a coloring book}} &
     \small \hl{\textit{Alex Hill, Original coloring book illustration, Moonlight Bay}} \\ \cline{2-4}
    & \small \textit{The great elf city of Rivendell, sitting atop a waterfall as cascades of water spill around it} &
     \small \hl{\textit{Add a giant red dragon}} &
     \small \hl{\textit{The great elf city of Rivendell, sitting atop a waterfall as cascades of water spill around it with a giant red dragon flying overhead}} \\ \cline{2-4}
    & \small \textit{Kate Hudson arriving at the Golden Globes 2015} &
     \small \hl{\textit{make her look like a zombie}} &
     \small \hl{\textit{Zombie Kate Hudson arriving at the Golden Globes 2015}} \\ \cline{2-4}
    & \small  {...} &
     \small  {...} &
     \small  {...} \\
    \bottomrule
  \end{tabular}
  }
  \caption{We label a small text dataset, finetune GPT-3, and use that finetuned model to generate a large dataset of text triplets. As the input caption for both the labeled and generated examples, we use real image captions from LAION. \hl{Highlighted text} is generated by GPT-3.}
  \label{tab:example}
\end{table*}

We first operate entirely in the text domain, where we leverage a large language model to take in image captions and produce editing instructions and the resulting text captions after the edit. For example, as shown in Figure~\ref{fig:method}a, provided the input caption \emph{``photograph of a girl riding a horse"}, our language model can generate both a plausible edit instruction \emph{``have her ride a dragon"} and an appropriately modified output caption \emph{``photograph of a girl riding a dragon"}. Operating in the text domain enables us to generate a large and diverse collection of edits, while maintaining correspondence between the image changes and text instructions.

Our model is trained by finetuning GPT-3 on a relatively small human-written dataset of editing triplets: (1) input captions, (2) edit instructions, (3) output captions. To produce the fine-tuning dataset, we sampled 700 input captions from the LAION-Aesthetics V2 6.5+ \cite{schuhmann2022laion} dataset and manually wrote instructions and output captions. See Table~\ref{tab:example}a for examples of our written instructions and output captions. Using this data, we fine-tuned the GPT-3 Davinci model for a single epoch using the default training parameters.

Benefiting from GPT-3's immense knowledge and ability to generalize, our finetuned model is able to generate creative yet sensible instructions and captions. See Table~\ref{tab:example}b for example GPT-3 generated data. Our dataset is created by generating a large number of edits and output captions using this trained model, where the input captions are real image captions from LAION-Aesthetics (excluding samples with duplicate captions or duplicate image URLs). We chose the LAION dataset due to its large size, diversity of content (including references to proper nouns and popular culture), and variety of mediums (photographs, paintings, digital artwork). A potential drawback of LAION is that it is quite noisy and contains a number of nonsensical or undescriptive captions---however, we found that dataset noise is mitigated through a combination of dataset filtering (Section~\ref{subsubsec:text2img}) and classifier-free guidance (Section~\ref{subsubsec:cfg}). Our final corpus of generated instructions and captions consists of $454,445$ examples.

\subsubsection{Generating Paired Images from Paired Captions}
\label{subsubsec:text2img}

\begin{figure}[t]
    \begin{subfigure}{0.49\linewidth}
        \includegraphics[width=0.495\linewidth]{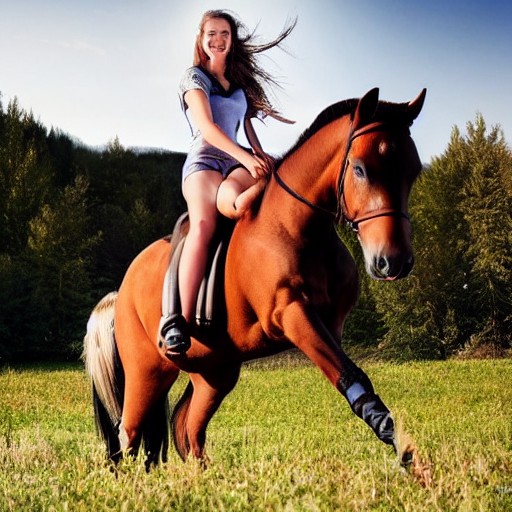}%
        \hfill
        \includegraphics[width=0.495\linewidth]{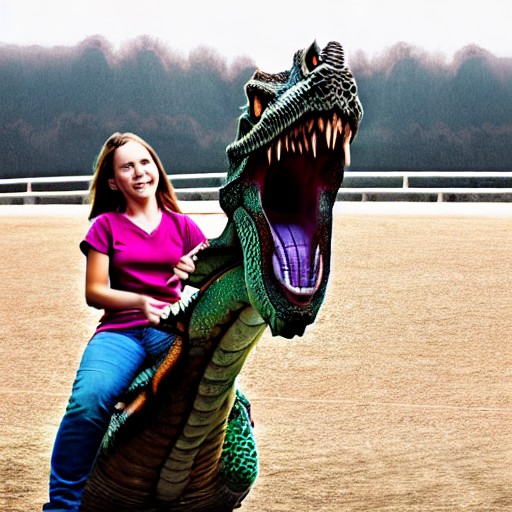}%
        \caption{Without Prompt-to-Prompt.}
    \end{subfigure}
    \hfill
    \begin{subfigure}{0.49\linewidth}
        \includegraphics[width=0.495\linewidth]{assets/p2p/horse_0.jpg}%
        \hfill
        \includegraphics[width=0.495\linewidth]{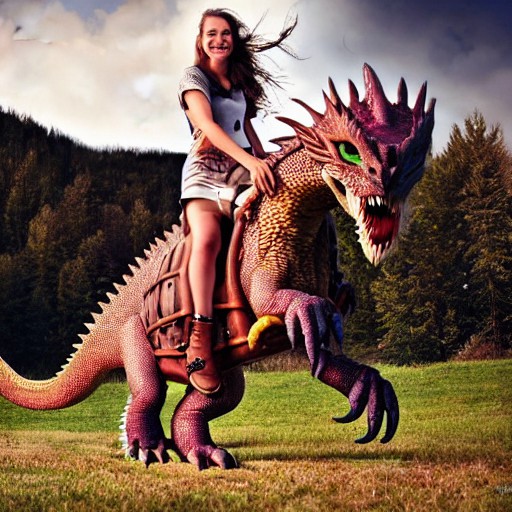}%
        \caption{With Prompt-to-Prompt.}
    \end{subfigure}
    \vspace{-.1in}
    \caption{Pair of images generated using StableDiffusion~\cite{rombach2022high} with and without Prompt-to-Prompt~\cite{hertz2022prompt}. For both, the corresponding captions are \textit{``photograph of a girl riding a horse"} and \textit{``photograph of a girl riding a dragon"}.}
    \label{fig:p2p}
\end{figure}

Next, we use a pretrained text-to-image model to transform a pair of captions (referring to the image before and after the edit) into a pair of images. 
One challenge in turning a pair of captions into a pair of corresponding images is that text-to-image models provide no guarantees about image consistency, even under very minor changes of the conditioning prompt. For example, two very similar prompts: \textit{``a picture of a cat''} and \textit{``a picture of a black cat''} may produce wildly different images of cats. This is unsuitable for our purposes, where we intend to use this paired data as supervision for training a model to edit images (and not produce a different random image). We therefore use Prompt-to-Prompt~\cite{hertz2022prompt}, a recent method aimed at encouraging multiple generations from a text-to-image diffusion model to be similar. This is done through borrowed cross attention weights in some number of denoising steps. Figure~\ref{fig:p2p} shows a comparison of sampled images with and without Prompt-to-Prompt. 

While this greatly helps assimilate generated images, different edits may require different amounts of change in image-space. For instance, changes of larger magnitude, such as those which change large-scale image structure (e.g., moving objects around, replacing with objects of different shapes), may require less similarity in the generated image pair. Fortunately, Prompt-to-Prompt has as a parameter that can control the similarity between the two images: the fraction of denoising steps $p$ with shared attention weights. Unfortunately, identifying an optimal value of $p$ from only the captions and edit text is difficult. We therefore generate $100$ sample pairs of images per caption-pair, each with a random $p \sim \mathcal{U}(0.1,0.9)$, and filter these samples by using a CLIP-based metric: the directional similarity in CLIP space as introduced by Gal \emph{et al.}~\cite{gal2022stylegan}. This metric measures the consistency of the change between the two images (in CLIP space) with the change between the two image captions. Performing this filtering not only helps maximize the diversity and quality of our image pairs, but also makes our data generation more robust to failures of Prompt-to-Prompt and Stable Diffusion. 

\subsection{InstructPix2Pix}
\label{subsec:model}

We use our generated training data to train a conditional diffusion model that edits images from written instructions.  We base our model on Stable Diffusion, a large-scale text-to-image latent diffusion model. 

Diffusion models~\cite{pmlrv37sohldickstein15} learn to generate data samples through a sequence of denoising autoencoders that estimate the score~\cite{hyvarinen2005estimation} of a data distribution (a direction pointing toward higher density data). Latent diffusion~\cite{rombach2022high} improves the efficiency and quality of diffusion models by operating in the latent space of a pretrained variational autoencoder~\cite{kingma2013auto} with encoder $\mathcal{E}$ and decoder $\mathcal{D}$. For an image $x$, the diffusion process adds noise to the encoded latent $z = \mathcal{E}(x)$ producing a noisy latent $z_t$ where the noise level increases over timesteps $t \in T$. We learn a network $\epsilon_\theta$ that predicts the noise added to the noisy latent $z_t$ given image conditioning $c_I$ and text instruction conditioning $c_T$. We minimize the following latent diffusion objective:

\vspace{-4mm}
\begin{equation}
L = \mathbb{E}_{\mathcal{E}(x), \mathcal{E}(c_I), c_T, \epsilon \sim \mathcal{N}(0, 1), t }\Big[ \Vert \epsilon - \epsilon_\theta(z_{t}, t, \mathcal{E}(c_I), c_T)) \Vert_{2}^{2}\Big]
\label{eq:loss}
\end{equation}
\vspace{-1mm}

Wang \emph{et al.}~\cite{wang2022pretraining} show that fine-tuning a large image diffusion models outperforms training a model from scratch for image translation tasks, especially when paired training data is limited. We therefore initialize the weights of our model with a pretrained Stable Diffusion checkpoint, leveraging its vast text-to-image generation capabilities. To support image conditioning, we add additional input channels to the first convolutional layer, concatenating $z_t$ and $\mathcal{E}(c_I)$. All available weights of the diffusion model are initialized from the pretrained checkpoints, and weights that operate on the newly added input channels are initialized to zero. We reuse the same text conditioning mechanism that was originally intended for captions to instead take as input the text edit instruction $c_T$. Additional training details are provided in the supplemental material.

\begin{figure}
    \renewcommand{\arraystretch}{0.3}
    \begin{tabular}{@{\hspace{0mm}}p{5.5mm}@{\hskip 0.75mm}c@{\hskip 0.4mm}c@{\hskip 0.4mm}c}
        \vspace{0.4mm} & 
        \scriptsize$s_T=3$ & \scriptsize$s_T=7.5$ & \scriptsize$s_T=15$ \\
        \vspace{-8mm}\vspace{-4mm}\scriptsize$s_I\hspace{-1mm}=\hspace{-1mm}1.0$\vspace{4mm} &%
        \includegraphics[width=.30\linewidth]{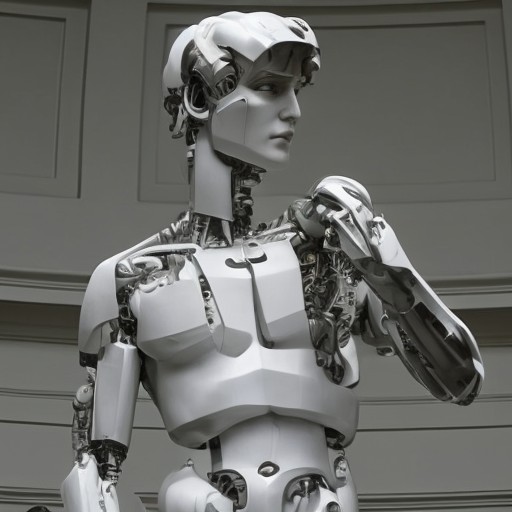} &%
        \includegraphics[width=.30\linewidth]{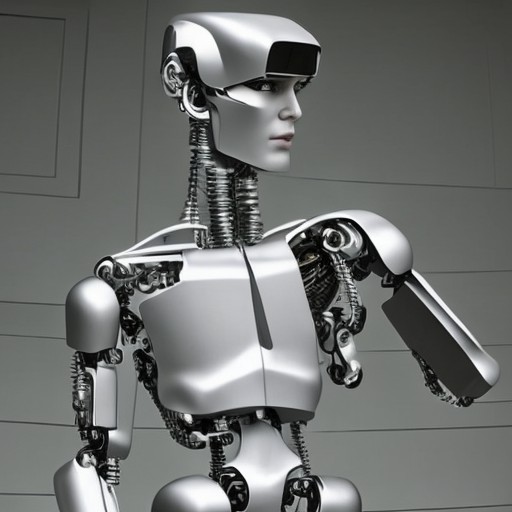} &%
        \includegraphics[width=.30\linewidth]{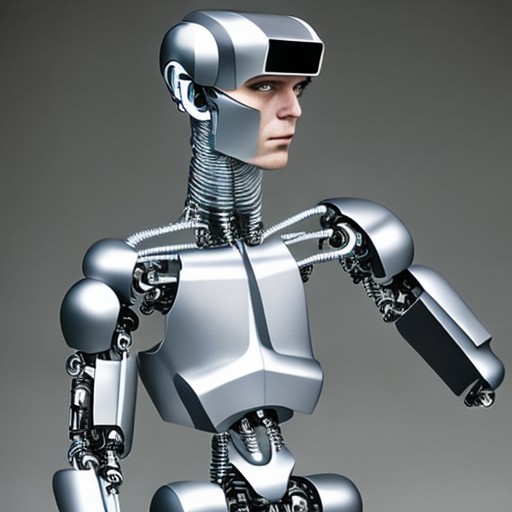}\\
        \vspace{-8mm}\vspace{-4mm}\scriptsize$s_I\hspace{-1mm}=\hspace{-1mm}1.2$\vspace{4mm} &%
        \includegraphics[width=.30\linewidth]{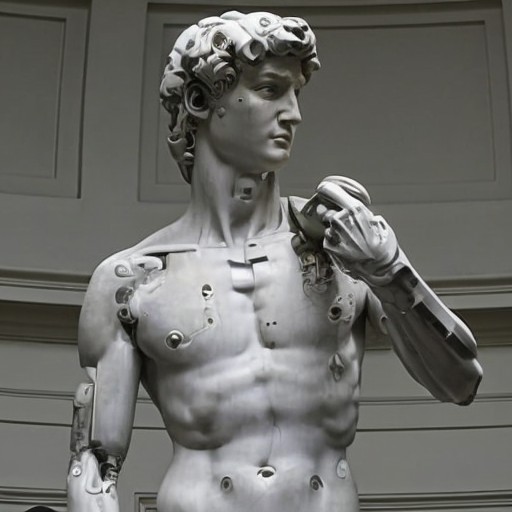} &%
        \includegraphics[width=.30\linewidth]{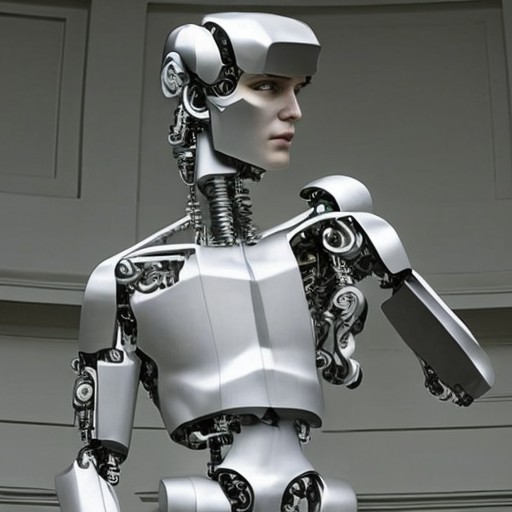} &%
        \includegraphics[width=.30\linewidth]{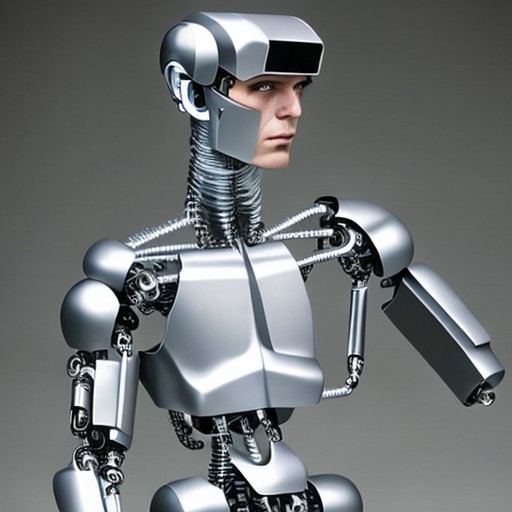}\\
        \vspace{-8mm}\vspace{-4mm}\scriptsize$s_I\hspace{-1mm}=\hspace{-1mm}1.6$\vspace{4mm} &%
        \includegraphics[width=.30\linewidth]{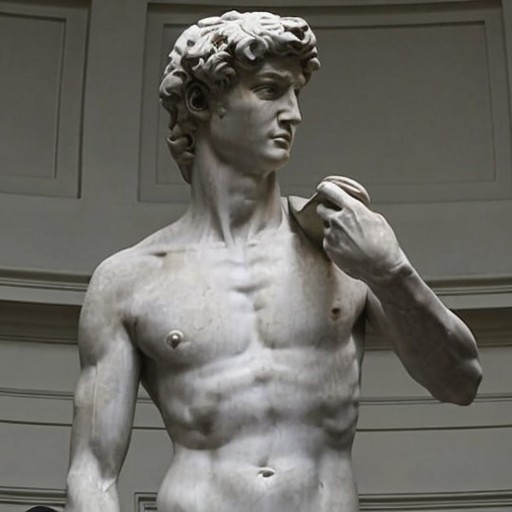} &%
        \includegraphics[width=.30\linewidth]{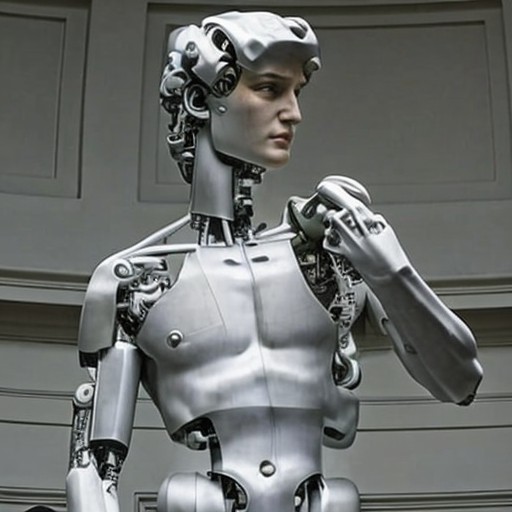} &%
        \includegraphics[width=.30\linewidth]{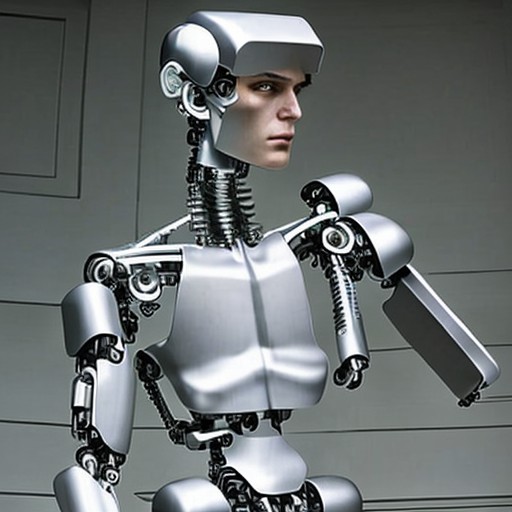}\\
    \end{tabular}
    \vspace{-4mm}
    \begin{center}\begin{scriptsize}Edit instruction: \emph{``Turn him into a cyborg!"}\end{scriptsize}\end{center}
    \vspace{-4mm}
    \caption{Classifier-free guidance weights over two conditional inputs. $s_I$ controls similarity with the input image, while $s_T$ controls consistency with the edit instruction. }
    \label{fig:cfg}
    \vspace{-1.5mm}
\end{figure} 

\subsubsection{Classifier-free Guidance for Two Conditionings}
\label{subsubsec:cfg}

Classifier-free diffusion guidance~\cite{ho2022classifier} is a method for trading off the quality and diversity of samples generated by a diffusion model. It is commonly used in class-conditional and text-conditional image generation to improve the visual quality of generated images and to make sampled images better correspond with their conditioning. Classifier-free guidance effectively shifts probability mass toward data where an implicit classifier $p_{\theta}(c|z_t)$ assigns high likelihood to the conditioning $c$. The implementation of classifier-free guidance involves jointly training the diffusion model for conditional and unconditional denoising, and combining the two score estimates at inference time. Training for unconditional denoising is done by simply setting the conditioning to a fixed null value $c\tight{=}\varnothing$ at some frequency during training. At inference time, with a guidance scale $s\ge1$, the modified score estimate $\tilde{e_{\theta}}(z_t, c)$ is extrapolated in the direction toward the conditional $e_{\theta}(z_t, c)$ and away from the unconditional $e_{\theta}(z_t, \varnothing)$.

\vspace{-3.5mm}
\begin{equation}
    \tilde{e_{\theta}}(z_t, c) = e_{\theta}(z_t, \varnothing) + s \cdot (e_{\theta}(z_t, c) - e_{\theta}(z_t, \varnothing))
    \label{eq:cfg}
\end{equation}
\vspace{-3mm}

For our task, the score network $e_{\theta}(z_t, c_I, c_T)$  has two conditionings: the input image $c_I$ and text instruction $c_T$. We find if beneficial to leverage classifier-free guidance with respect to both conditionings. Liu \emph{et al.}~\cite{liu2022compositional} demonstrate that a conditional diffusion model can compose score estimates from multiple different conditioning values.
We apply the same concept to our model with two separate conditioning inputs. During training, we randomly set only $c_I\tight{=}\varnothing_I$ for 5\% of examples, only $c_T\tight{=}\varnothing_T$ for 5\% of examples, and both $c_I\tight{=}\varnothing_I$ and $c_T\tight{=}\varnothing_T$ for 5\% of examples. Our model is therefore capable of conditional or unconditional denoising with respect to both or either conditional inputs. We introduce two guidance scales, $s_I$ and $s_T$, which can be adjusted to trade off how strongly the generated samples correspond with the input image and how strongly they correspond with the edit instruction.
Our modified score estimate is as follows: 

\vspace{-3.5mm}
\begin{equation}
\begin{split}
    \tilde{e_{\theta}}(z_t, c_I, c_T) = &\: e_{\theta}(z_t, \varnothing, \varnothing) \\ &+ s_I \cdot (e_{\theta}(z_t, c_I, \varnothing) - e_{\theta}(z_t, \varnothing, \varnothing)) \\ &+ s_T \cdot (e_{\theta}(z_t, c_I, c_T) - e_{\theta}(z_t, c_I, \varnothing))
    \label{eq:cfg2}
\end{split}
\end{equation}

In Figure~\ref{fig:cfg}, we show the effects of these two parameters on generated samples. See Appendix~\ref{sec:cfg2a} for details of our classifier-free guidance formulation.

\section{Results}
\label{sec:results}

\begin{figure*}[h!]
    \renewcommand{\arraystretch}{0.3}
    \setlength\tabcolsep{1.5pt}
    \resizebox{\linewidth}{!}{%
    \begin{tabular}{ccccc}
        \includegraphics[width=0.205\textwidth]{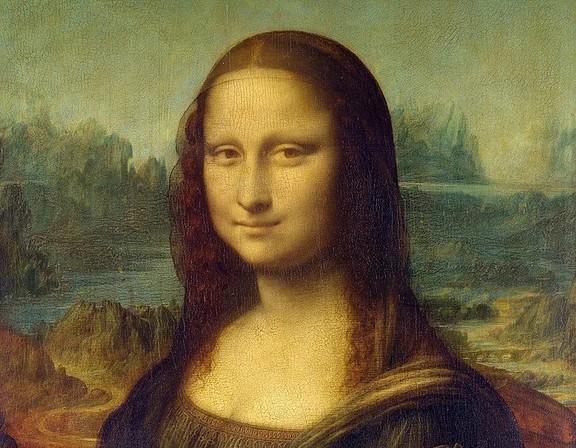} &%
        \includegraphics[width=0.205\textwidth]{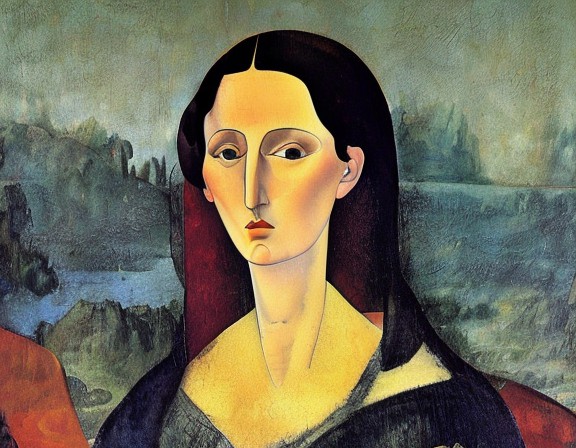} &%
        \includegraphics[width=0.205\textwidth]{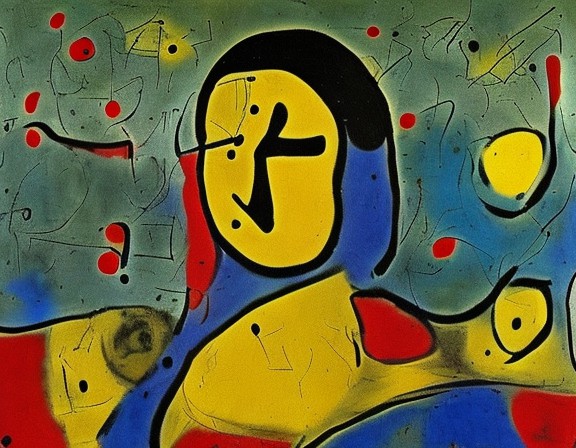} &%
        \includegraphics[width=0.205\textwidth]{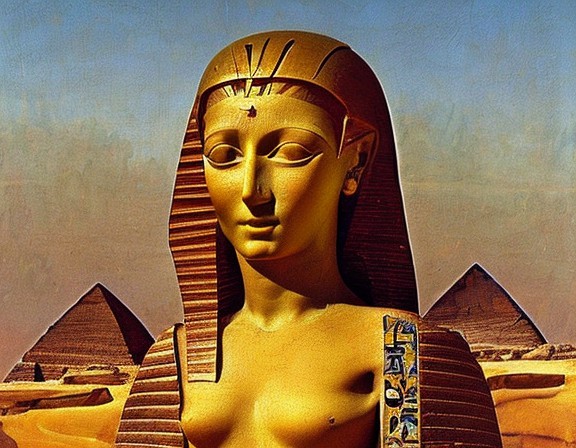} &%
        \includegraphics[width=0.205\textwidth]{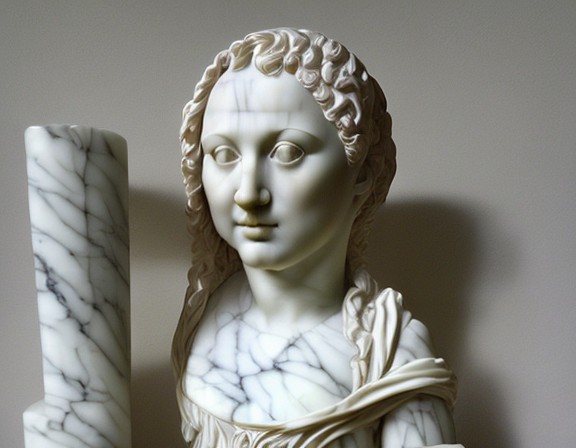}\\
         \scriptsize{Input} & \scriptsize{\emph{``Make it a Modigliani painting''}} & \scriptsize{\emph{``Make it a Miro painting''}} & \scriptsize{\emph{``Make it an Egyptian sculpture''}} & \scriptsize{\emph{``Make it a marble roman sculpture''}} \\ \\
    \end{tabular}}
    \vspace{-3mm}
    \caption{\emph{Mona Lisa} transformed into various artistic mediums.}
    \label{fig:mona_lisa}
\end{figure*}

\begin{figure*}[h!]
    \renewcommand{\arraystretch}{0.3}
    \setlength\tabcolsep{1.5pt}
    \resizebox{\linewidth}{!}{%
    \begin{tabular}{ccc}
        \includegraphics[width=0.334\textwidth]{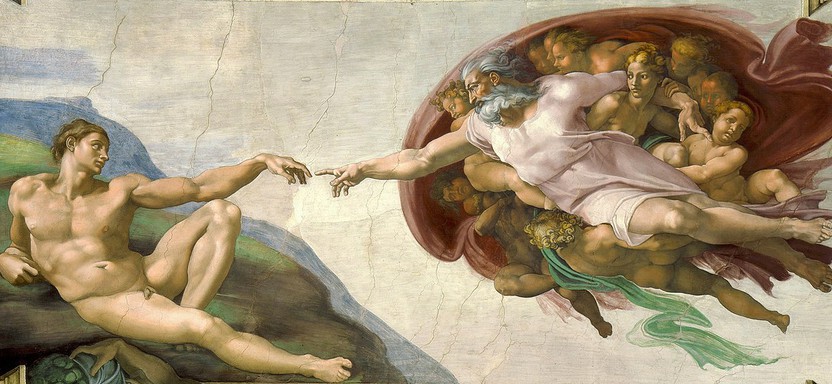} &%
        \includegraphics[width=0.334\textwidth]{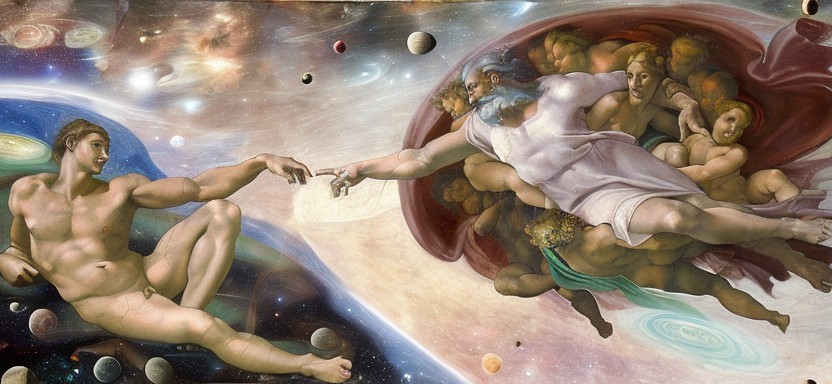} &%
        \includegraphics[width=0.334\textwidth]{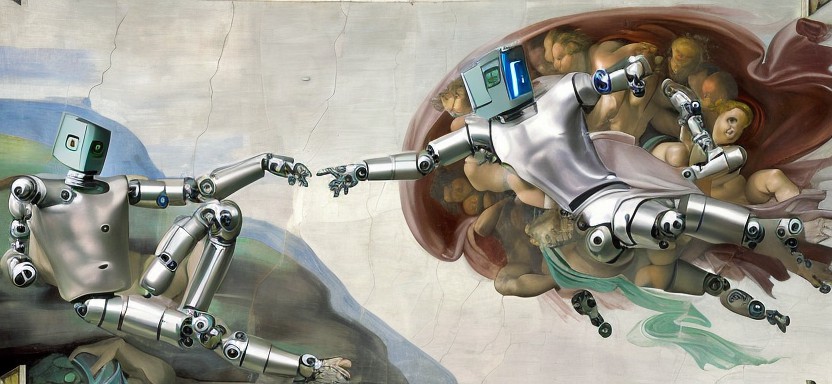} \\
          \scriptsize{Input} & \scriptsize{\emph{``Put them in outer space''}} & \scriptsize{\emph{``Turn the humans into robots''}} \\ \\
    \end{tabular}}
    \vspace{-3mm}
    \caption{\emph{The Creation of Adam} with new context and subjects (generated at 768 resolution).}
    \label{fig:adam}
\end{figure*}

\begin{figure*}[h!]
    \renewcommand{\arraystretch}{0.3}
    \setlength\tabcolsep{1.5pt}
    \resizebox{\linewidth}{!}{%
    \begin{tabular}{cccc}
        \includegraphics[width=0.26\textwidth]{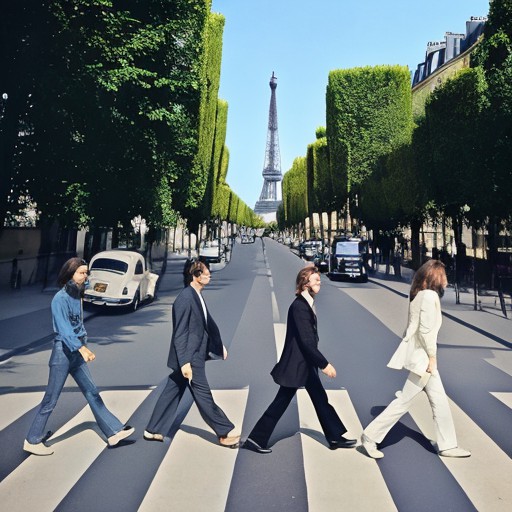} &%
        \includegraphics[width=0.26\textwidth]{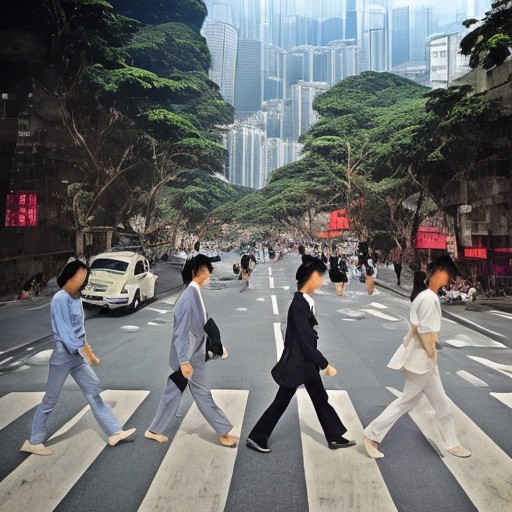} &%
        \includegraphics[width=0.26\textwidth]{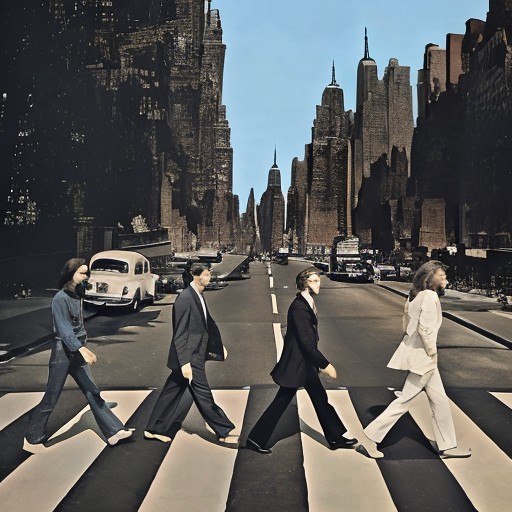} &%
        \includegraphics[width=0.26\textwidth]{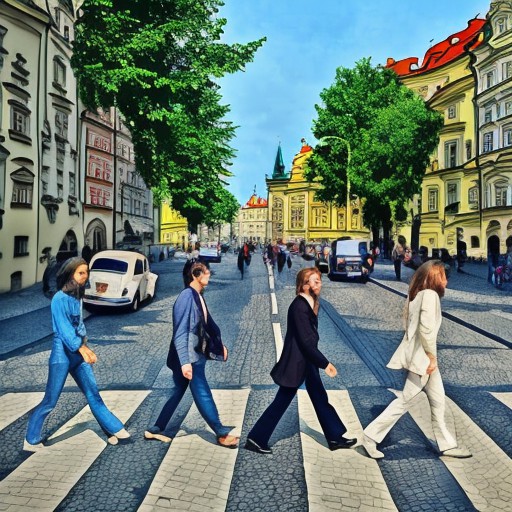} \\
        \scriptsize{\emph{``Make it Paris''}} & \scriptsize{\emph{``Make it Hong Kong''}} & \scriptsize{\emph{``Make it Manhattan''}} & \scriptsize{\emph{``Make it Prague''}} \\ \\
        \includegraphics[width=0.26\textwidth]{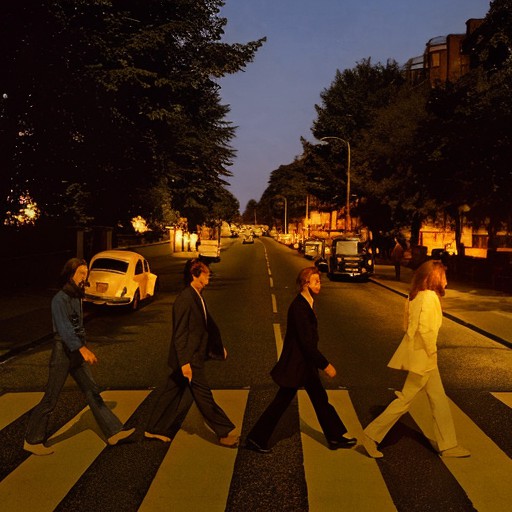} &%
        \includegraphics[width=0.26\textwidth]{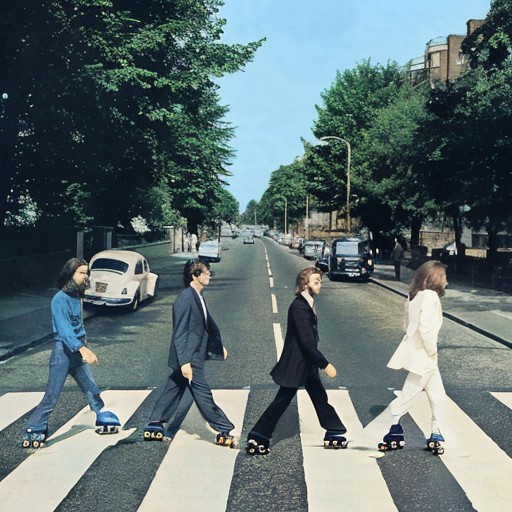} &%
        \includegraphics[width=0.26\textwidth]{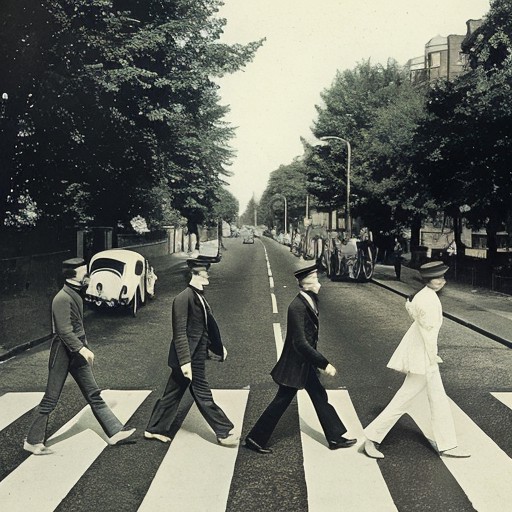} &%
        \includegraphics[width=0.26\textwidth]{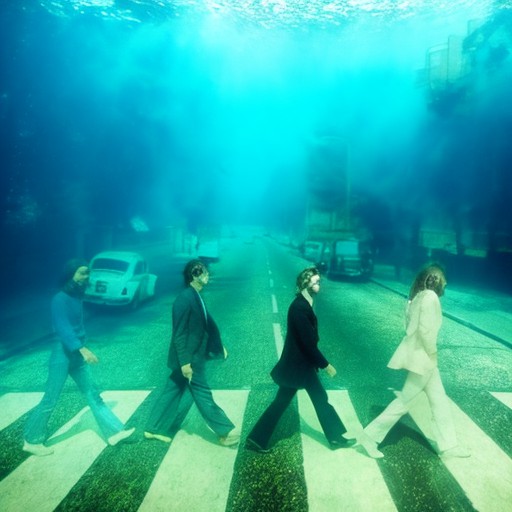}\\
        \scriptsize{\emph{``Make it evening''}} & \scriptsize{\emph{``Put them on roller skates''}} & \scriptsize{\emph{``Turn this into 1900s''}} & \scriptsize{\emph{``Make it underwater''}} \\ \\
        \includegraphics[width=0.26\textwidth]{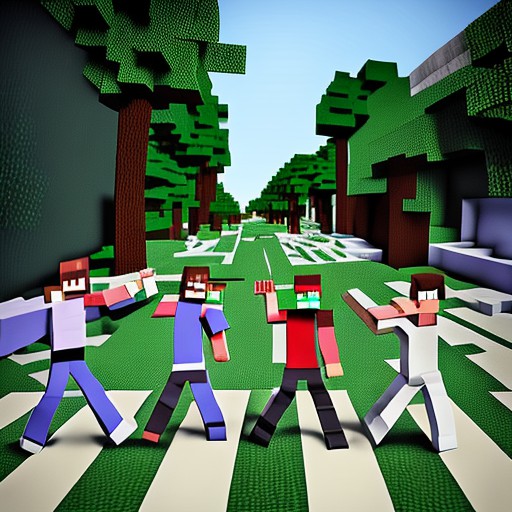} &%
        \includegraphics[width=0.26\textwidth]{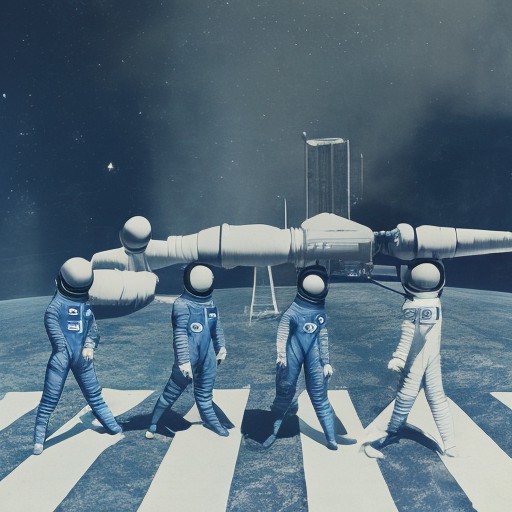} &%
        \includegraphics[width=0.26\textwidth]{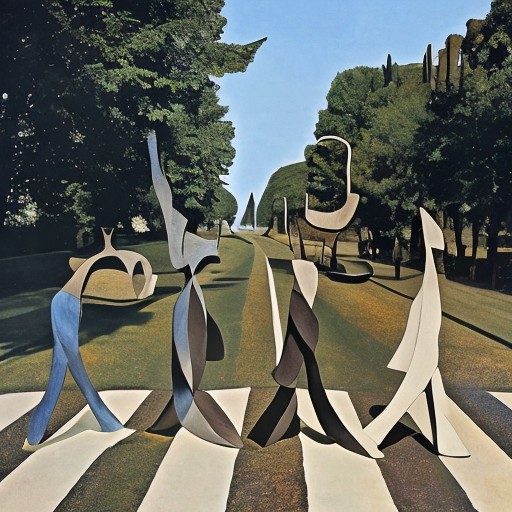} &%
        \includegraphics[width=0.26\textwidth]{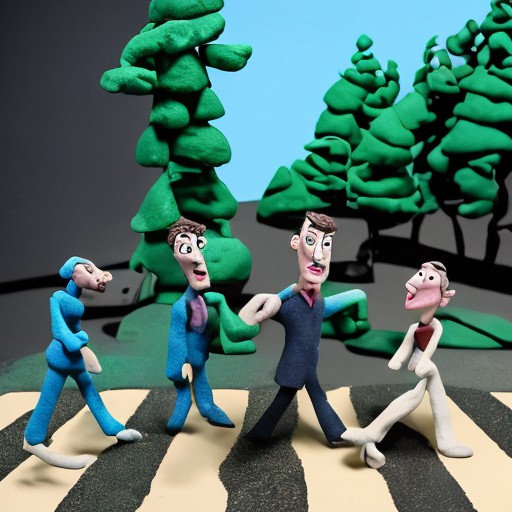} \\
        \scriptsize{\emph{``Make it Minecraft''}} & \scriptsize{\emph{``Turn this into the space age''}} & \scriptsize{\emph{``Make them into Alexander Calder sculptures''}} & \scriptsize{\emph{``Make it a Claymation''}} \\ \\
    \end{tabular}}
    \vspace{-3mm}
    \caption{The iconic Beatles \emph{Abbey Road} album cover transformed in a variety of ways.}
    \label{fig:beatles}
\end{figure*}

\begin{figure*}[t]
    \renewcommand{\arraystretch}{0.3}
    \includegraphics[width=\textwidth]{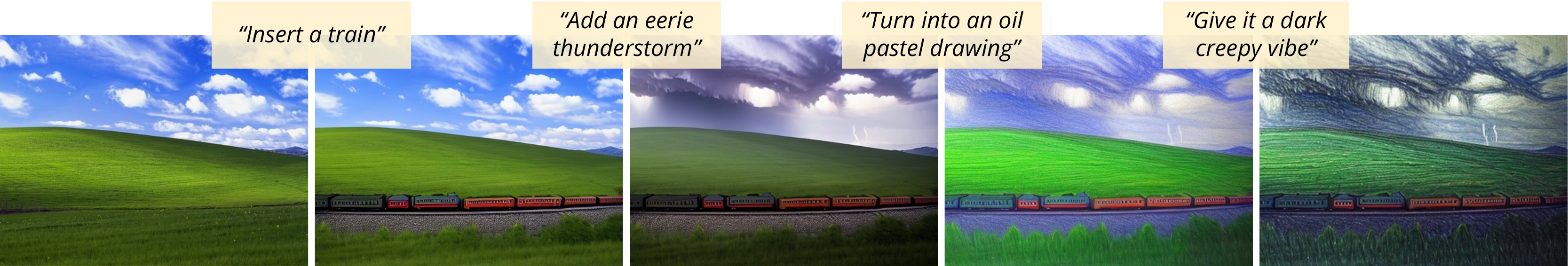}
    \vspace{-.2in}
    \caption{Applying our model recurrently with different instructions results in compounded edits.}
    \label{fig:chained_edits}
    \vspace{1mm}
\end{figure*}

\begin{figure*}[t]
    \begin{subfigure}{0.16\linewidth}
        \includegraphics[width=\linewidth]{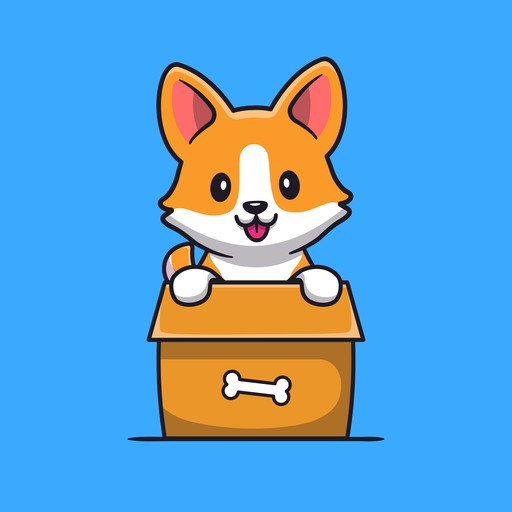}
        \begin{center}
        \vspace{-2.5mm}
        \scriptsize Input
        \vspace{-1.5mm}
        \end{center}
    \end{subfigure}
    \hfill
    \begin{subfigure}{0.82\linewidth}
        \includegraphics[width=0.1951\linewidth]{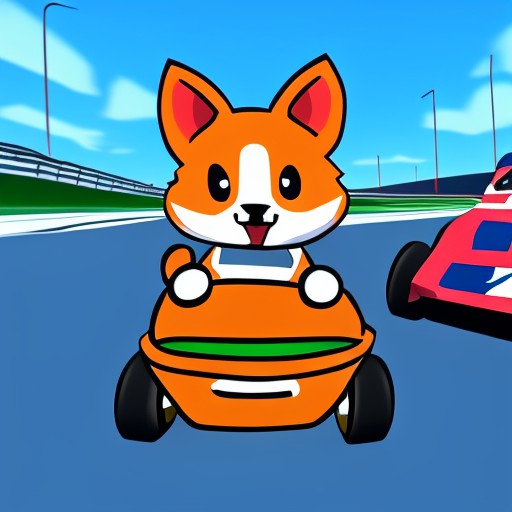}%
        \hfill%
        \includegraphics[width=0.1951\linewidth]{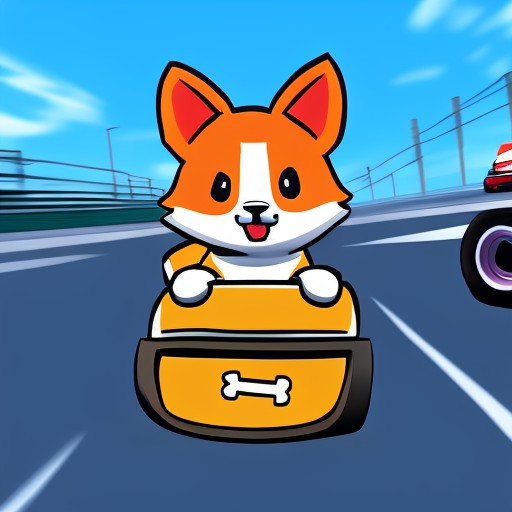}%
        \hfill%
        \includegraphics[width=0.1951\linewidth]{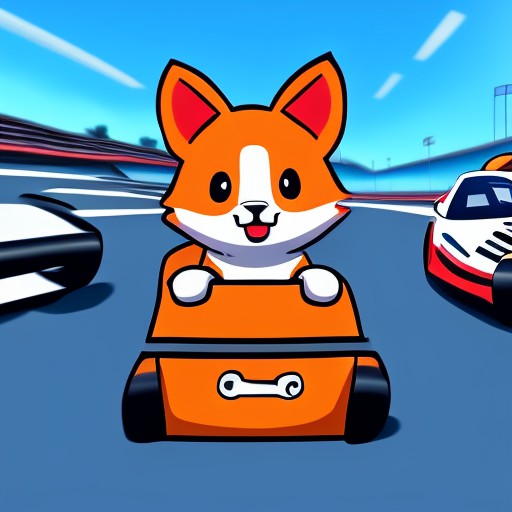}%
        \hfill%
        \includegraphics[width=0.1951\linewidth]{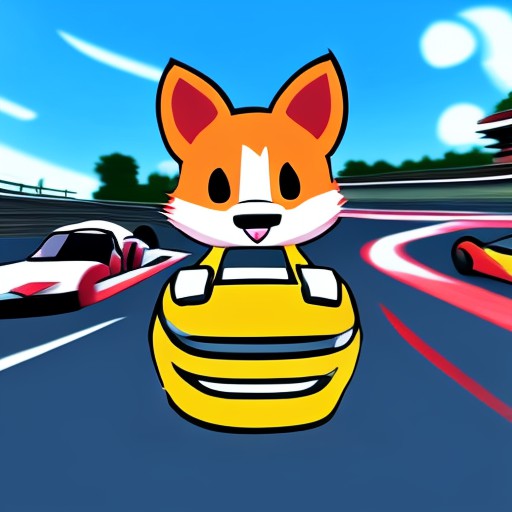}%
        \hfill%
        \includegraphics[width=0.1951\linewidth]{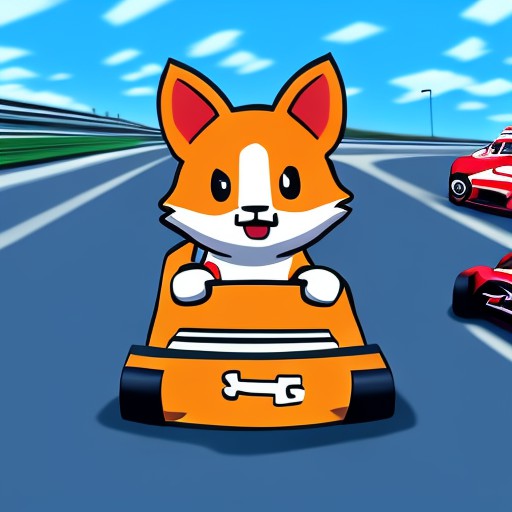}%
        \begin{center}
        \vspace{-2.5mm}
        \scriptsize Edit instruction: \emph{``in a race car video game"}
        \vspace{-1.5mm}
        \end{center}
    \end{subfigure}
    \vspace{-.5mm}
    \caption{By varying the latent noise, our model can produce many possible image edits for the same input image and instruction.}
    \label{fig:varying}
    \vspace{1mm}
\end{figure*}

\begin{figure*}[h!]
    \centering
    \begin{subfigure}[t]{0.496\linewidth}
        \includegraphics[width=\linewidth]{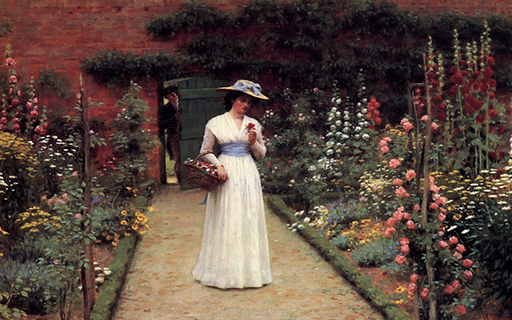}%
        \begin{center}
        \vspace{-2.5mm}
        \scriptsize Input
        \vspace{-1.5mm}
        \end{center}
    \end{subfigure}%
    \hfill
    \begin{subfigure}[t]{0.496\linewidth}
        \includegraphics[width=\linewidth]{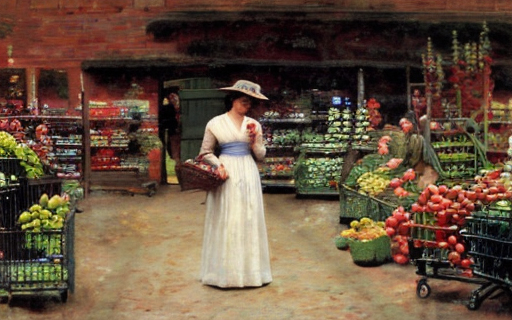}%
        \begin{center}
        \vspace{-2.5mm}
        \scriptsize \emph{``Make it a grocery store"}
        \vspace{-1.5mm}
        \end{center}
    \end{subfigure}%
    \caption{Leighton's \emph{Lady in a Garden} moved to a new setting.}
    \vspace{1mm}
    \label{fig:garden}
\end{figure*}

\begin{figure*}[h!]
    \begin{subfigure}[t]{0.327\linewidth}
        \includegraphics[width=\linewidth]{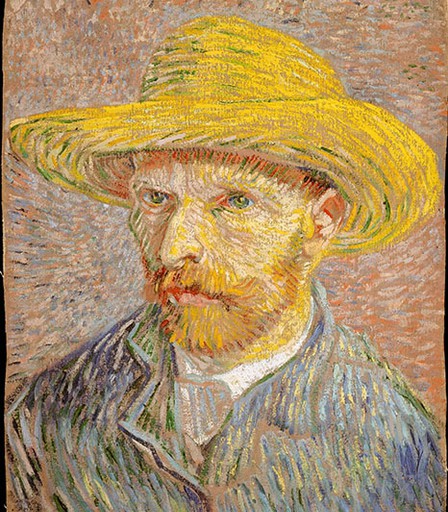}%
        \begin{center}
        \vspace{-2.5mm}
        \scriptsize Input
        \vspace{-1.5mm}
        \end{center}
    \end{subfigure}%
    \hfill%
    \begin{subfigure}[t]{0.327\linewidth}
        \includegraphics[width=\linewidth]{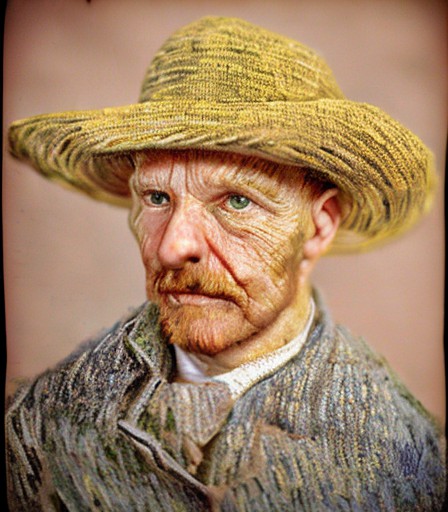}
        \begin{center}
        \vspace{-2.5mm}
        \scriptsize \emph{``Convert to a realistic photo"}
        \vspace{-1.5mm}
        \end{center}
    \end{subfigure}%
    \hfill
     \begin{subfigure}[t]{0.327\linewidth}
        \includegraphics[width=\linewidth]{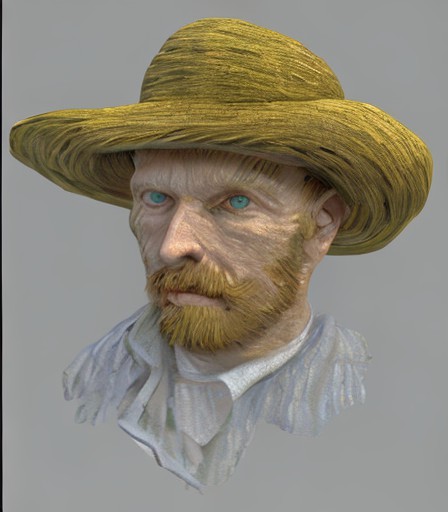}
        \begin{center}
        \vspace{-2.5mm}
        \scriptsize \emph{``Turn into a 3D model"}
        \vspace{-1.5mm}
        \end{center}
    \end{subfigure}%
    \caption{Van Gogh's \emph{Self-Portrait with a Straw Hat} in different mediums.}
    \vspace{1mm}
    \label{fig:vangogh}
\end{figure*}

\begin{figure*}[h!]
    \begin{subfigure}[t]{0.327\linewidth}
        \includegraphics[width=\linewidth]{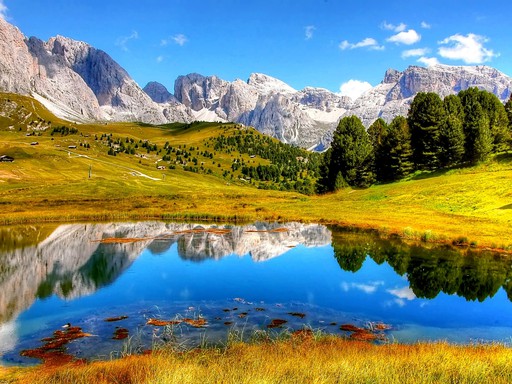}%
        \begin{center}
        \vspace{-2.5mm}
        \scriptsize Input
        \vspace{-1.5mm}
        \end{center}
    \end{subfigure}%
    \hfill%
    \begin{subfigure}[t]{0.327\linewidth}
        \includegraphics[width=\linewidth]{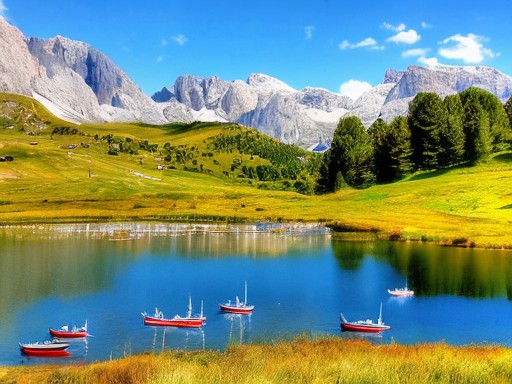}
        \begin{center}
        \vspace{-2.5mm}
        \scriptsize \emph{``Add boats on the water"}
        \vspace{-1.5mm}
        \end{center}
    \end{subfigure}%
    \hfill%
     \begin{subfigure}[t]{0.327\linewidth}
        \includegraphics[width=\linewidth]{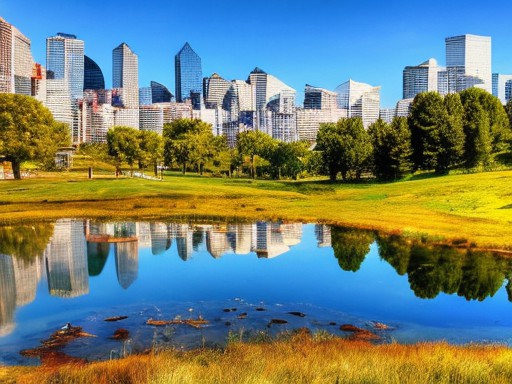}
        \begin{center}
        \vspace{-2.5mm}
        \scriptsize \emph{``Replace the mountains with a city skyline"}
        \vspace{-1.5mm}
        \end{center}
    \end{subfigure}%
    \caption{A landscape photograph shown with different contextual edits. Note that isolated changes also bring along accompanying contextual effects: the addition of boats also adds wind ripples in the water, and the added city skyline is reflected on the lake.}
    \vspace{0.5mm}
    \label{fig:landscape}
\end{figure*}

\begin{figure*}[h!]
    \begin{subfigure}[t]{0.327\linewidth}
        \includegraphics[width=\linewidth]{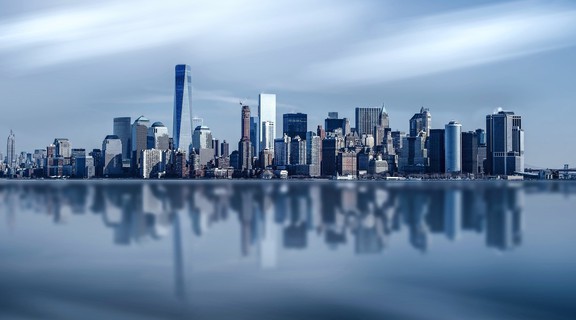}%
        \begin{center}
        \vspace{-2.5mm}
        \scriptsize Input
        \vspace{-1.5mm}
        \end{center}
    \end{subfigure}%
    \hfill%
    \begin{subfigure}[t]{0.327\linewidth}
        \includegraphics[width=\linewidth]{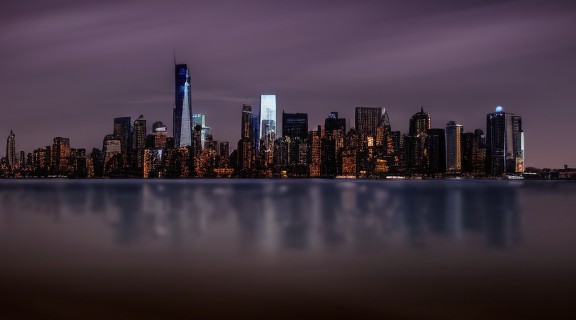}
        \begin{center}
        \vspace{-2.5mm}
        \scriptsize \emph{``It is now midnight"}
        \vspace{-1.5mm}
        \end{center}
    \end{subfigure}%
    \hfill%
     \begin{subfigure}[t]{0.327\linewidth}
        \includegraphics[width=\linewidth]{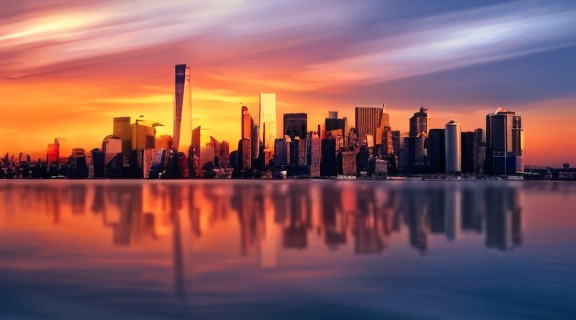}
        \begin{center}
        \vspace{-2.5mm}
        \scriptsize \emph{``Add a beautiful sunset"}
        \vspace{-1.5mm}
        \end{center}
    \end{subfigure}%
    \caption{A photograph of a cityscape edited to show different times of day.}
    \vspace{1mm}
    \label{fig:cityscape}
\end{figure*}

\begin{figure*}[h!]
    \begin{subfigure}[t]{0.195\linewidth}
        \includegraphics[trim={3.5cm 1cm 0 0},clip,width=\linewidth]{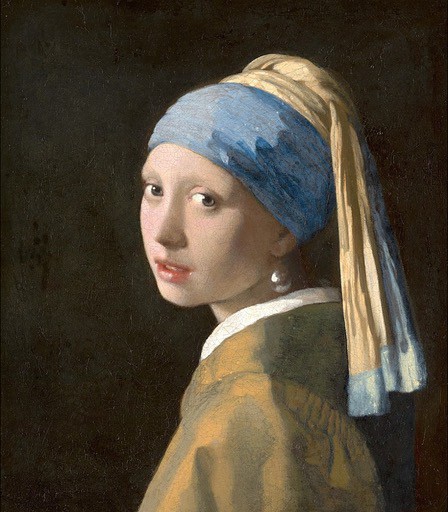}%
        \begin{center}
        \vspace{-2.5mm}
        \scriptsize Input
        \vspace{-1.5mm}
        \end{center}
    \end{subfigure}%
    \hfill%
    \begin{subfigure}[t]{0.195\linewidth}
        \includegraphics[trim={3.5cm 1cm 0 0},clip,width=\linewidth]{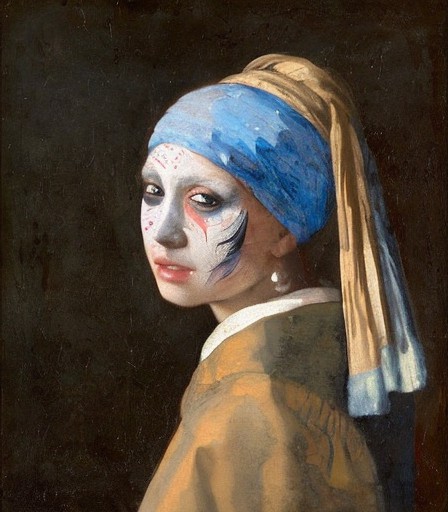}%
        \begin{center}
        \vspace{-2.5mm}
        \scriptsize \emph{``Apply face paint"}
        \vspace{-1.5mm}
        \end{center}
    \end{subfigure}%
    \hfill%
    \begin{subfigure}[t]{0.195\linewidth}
        \includegraphics[trim={3.5cm 1cm 0 0},clip,width=\linewidth]{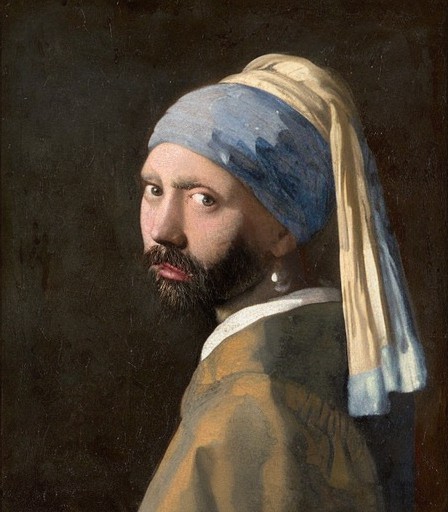}%
        \begin{center}
        \vspace{-2.5mm}
        \scriptsize \emph{``What would she look like as a bearded man?"}
        \vspace{-1.5mm}
        \end{center}
    \end{subfigure}%
    \hfill%
    \begin{subfigure}[t]{0.195\linewidth}
        \includegraphics[trim={3.5cm 1cm 0 0},clip,width=\linewidth]{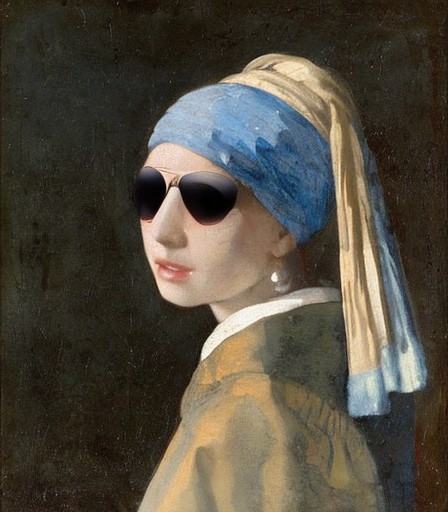}%
        \begin{center}
        \vspace{-2.5mm}
        \scriptsize \emph{``Put on a pair of sunglasses"}
        \vspace{-1.5mm}
        \end{center}
    \end{subfigure}%
    \hfill%
    \begin{subfigure}[t]{0.195\linewidth}
        \includegraphics[trim={3.5cm 1cm 0 0},clip,width=\linewidth]{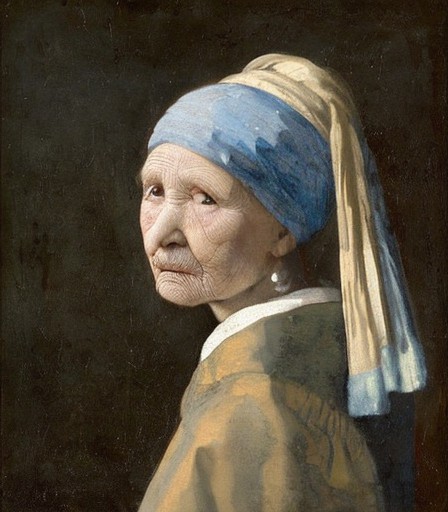}%
        \begin{center}
        \vspace{-2.5mm}
        \scriptsize \emph{``She should look 100 years old"}
        \vspace{-1.5mm}
        \end{center}
    \end{subfigure}%
    \vspace{3.5mm}
    \\ 
        \begin{subfigure}[t]{0.195\linewidth}
        \includegraphics[trim={3.5cm 1cm 0 0},clip,width=\linewidth]{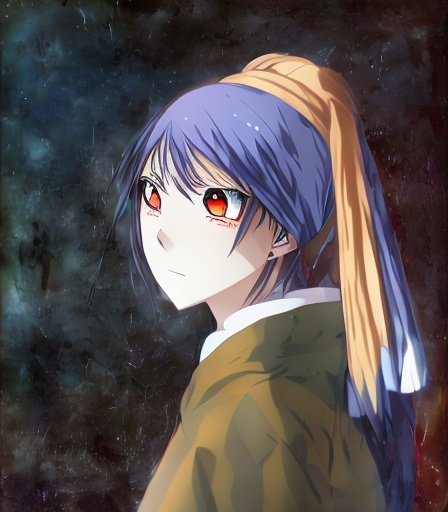}%
        \begin{center}
        \vspace{-2.5mm}
        \scriptsize \emph{``What if she were in an anime?"}
        \vspace{-1.5mm}
        \end{center}
    \end{subfigure}%
    \hfill%
    \begin{subfigure}[t]{0.195\linewidth}
        \includegraphics[trim={3.5cm 1cm 0 0},clip,width=\linewidth]{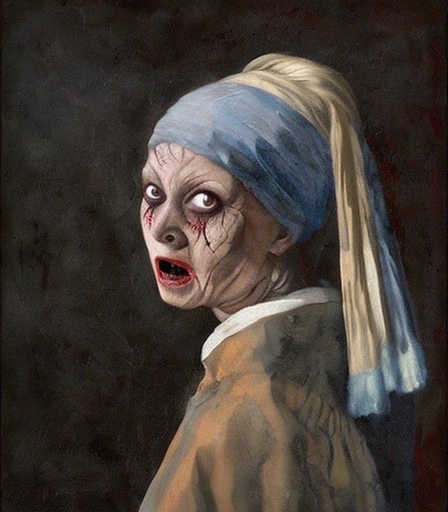}%
        \begin{center}
        \vspace{-2.5mm}
        \scriptsize \emph{``Make her terrifying"}
        \vspace{-1.5mm}
        \end{center}
    \end{subfigure}%
    \hfill%
    \begin{subfigure}[t]{0.195\linewidth}
        \includegraphics[trim={3.5cm 1cm 0 0},clip,width=\linewidth]{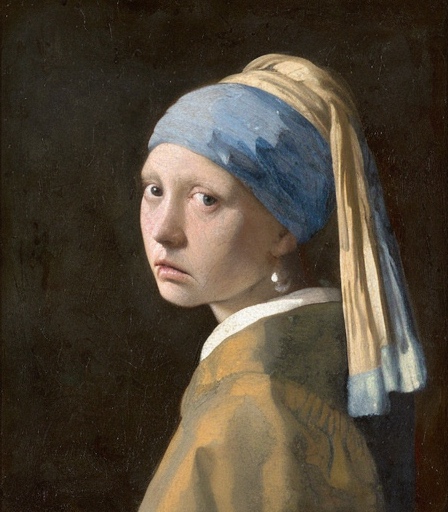}%
        \begin{center}
        \vspace{-2.5mm}
        \scriptsize \emph{``Make her more sad"}
        \vspace{-1.5mm}
        \end{center}
    \end{subfigure}%
    \hfill%
    \begin{subfigure}[t]{0.195\linewidth}
        \includegraphics[trim={3.5cm 1cm 0 0},clip,width=\linewidth]{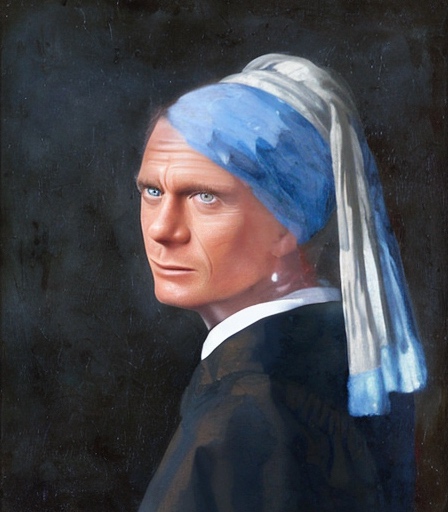}%
        \begin{center}
        \vspace{-2.5mm}
        \scriptsize \emph{``Make her James Bond"}
        \vspace{-1.5mm}
        \end{center}
    \end{subfigure}%
    \hfill%
    \begin{subfigure}[t]{0.195\linewidth}
        \includegraphics[trim={3.5cm 1cm 0 0},clip,width=\linewidth]{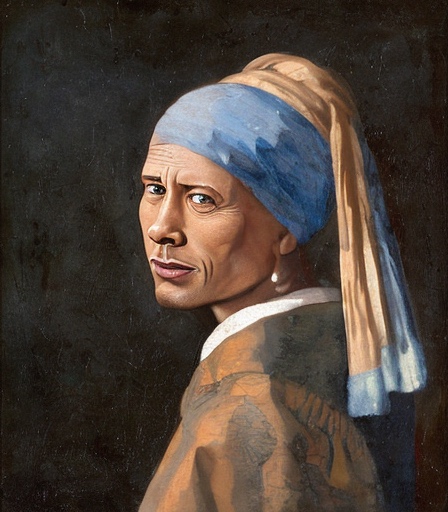}%
        \begin{center}
        \vspace{-2.5mm}
        \scriptsize \emph{``Turn her into Dwayne The Rock Johnson"}
        \vspace{-1.4mm}
        \end{center}
    \end{subfigure}%
    \vspace{-1.5mm}
    \caption{Vermeer's \emph{Girl with a Pearl Earring} with a variety of edits.}
    \label{fig:pearl}
\end{figure*}

We show instruction-based image editing results on a diverse set of real photographs and artwork, for a variety of types of edits and instruction wordings. See Figures~\ref{fig:teaser},~\ref{fig:mona_lisa},~\ref{fig:adam},~\ref{fig:beatles},~\ref{fig:chained_edits},~\ref{fig:varying},~\ref{fig:garden},~\ref{fig:vangogh},~\ref{fig:landscape},~\ref{fig:cityscape},~and~\ref{fig:pearl} for selected results. Our model successfully performs many challenging edits, including replacing objects, changing seasons and weather, replacing backgrounds, modifying material attributes, converting artistic medium, and a variety of others. 

We compare our method qualitatively with a couple recent works, SDEdit~\cite{meng2021sdedit} and Text2Live~\cite{bar2022text2live}. Our model follows instructions for how to edit the image, but prior works (including these baseline methods) expect descriptions of the image (or edit layer). Therefore, we provide them with the ``after-edit" text caption instead of the edit instruction. 
We also compare our method quantitatively with SDEdit, using two metrics measuring image consistency and edit quality, further described in Section~\ref{subsec:baselines}. Finally, we show ablations on how the size and quality of generated training data affect our model's performance in Section~\ref{sec:ablations}.

\begin{figure}[t]
    \centering
    \includegraphics[width=\linewidth]{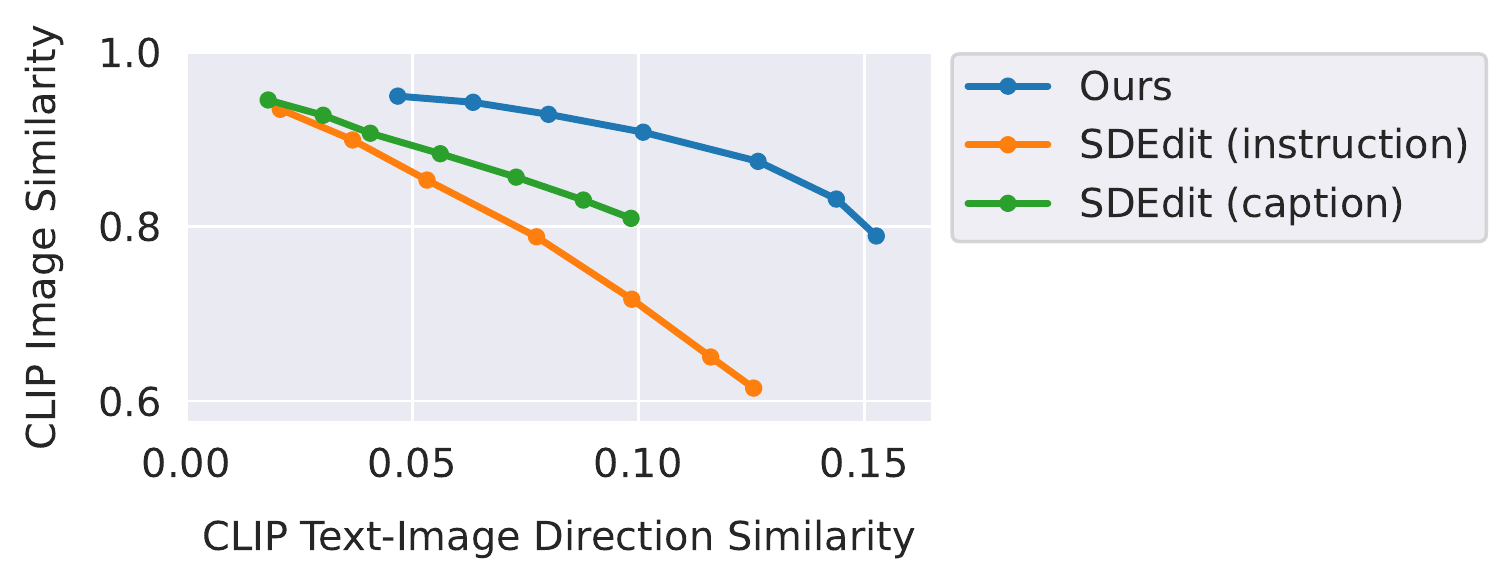}%
    \vspace{-.1in}
    \caption{We plot the trade-off between consistency with the input image (Y-axis) and consistency with the edit (X-axis). For both metrics, higher is better. For both methods, we fix text guidance to 7.5, and vary our $s_I\in[1.0,2.2]$ and SDEdit's strength (the amount of denoising) between $[0.3,0.9]$.}
    \label{fig:baselines}
\end{figure}

\begin{figure}[t]
    \renewcommand{\arraystretch}{0.5}
    \setlength\tabcolsep{1.5pt}
    \begin{tabular}{cc|cccc|ccc}
    \scriptsize Input & & & \scriptsize SDEdit-OC \cite{meng2021sdedit} & \scriptsize T2L \cite{bar2022text2live} & & & \scriptsize SDEdit-E \cite{meng2021sdedit} & \scriptsize \textbf{Ours}  \\
        \includegraphics[width=.18\linewidth]{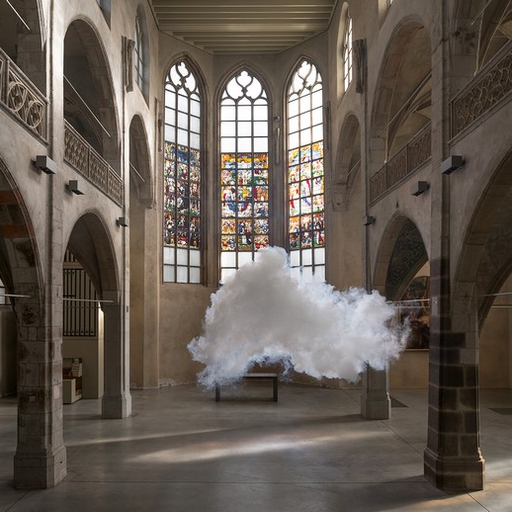} & & & %
        \multicolumn{2}{c}{\includegraphics[width=.18\linewidth]{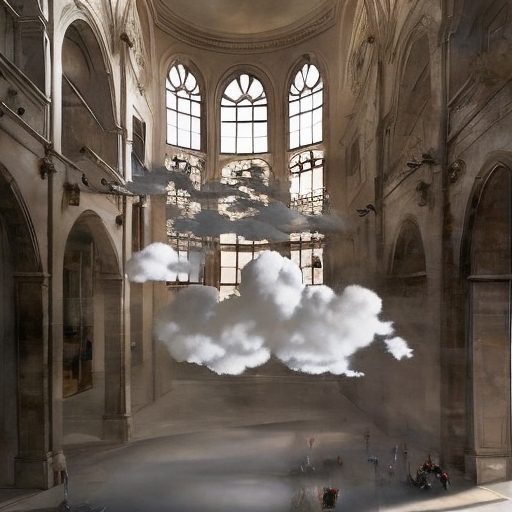} \includegraphics[width=.18\linewidth]{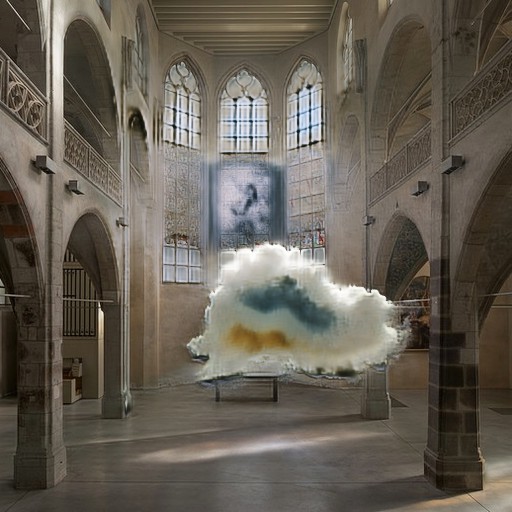}} & & & 
        \multicolumn{2}{c}{\includegraphics[width=.18\linewidth]{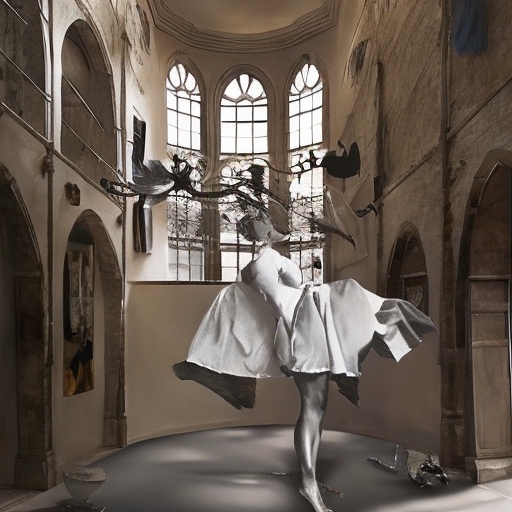} \includegraphics[width=.18\linewidth]{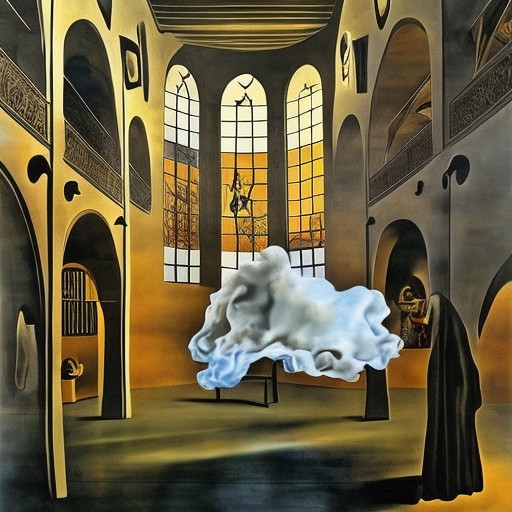}} \\
            & & & \multicolumn{2}{c}{\tiny{``Dali Painting of Nimbus Cloud..."}} & & &\multicolumn{2}{c}{\tiny{``make it look like a Dali Painting"}}\\
        \includegraphics[width=.18\linewidth]{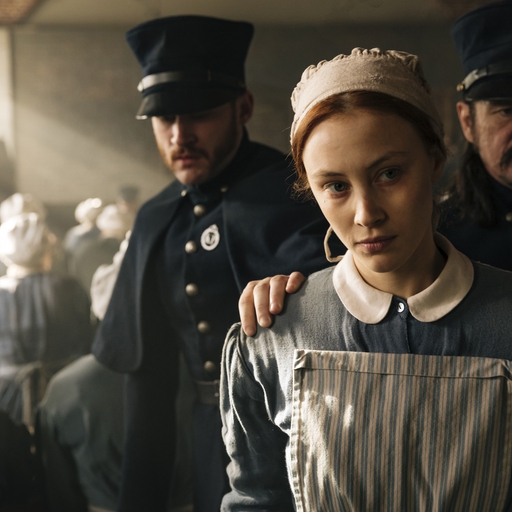} & & &%
        \multicolumn{2}{c}{\includegraphics[width=.18\linewidth]{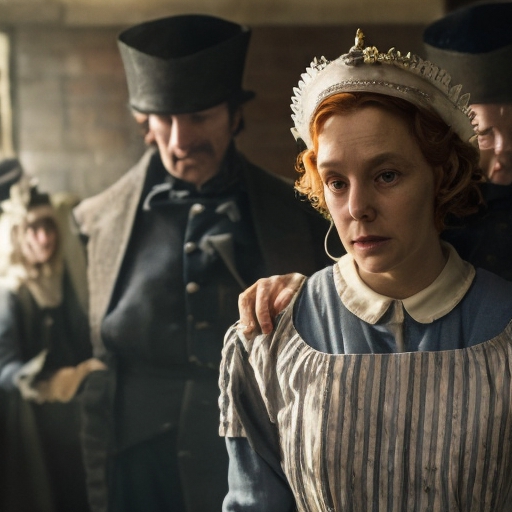} \includegraphics[width=.18\linewidth]{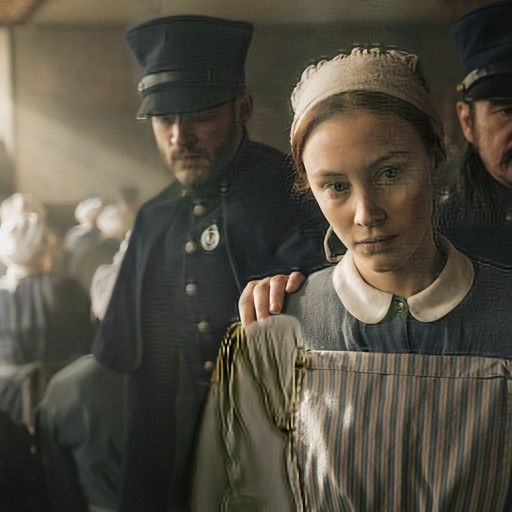}} & & & 
        \multicolumn{2}{c}{\includegraphics[width=.18\linewidth]{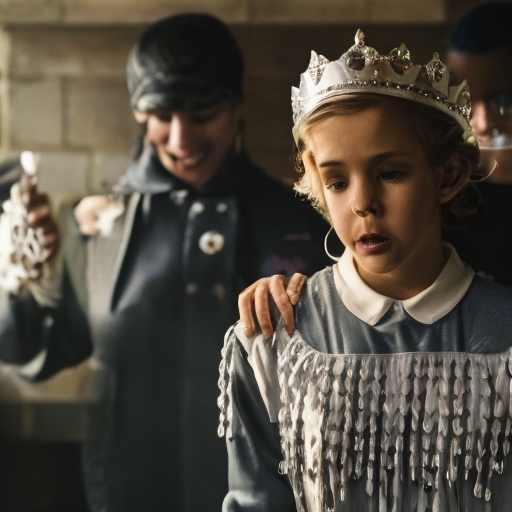}
        \includegraphics[width=.18\linewidth]{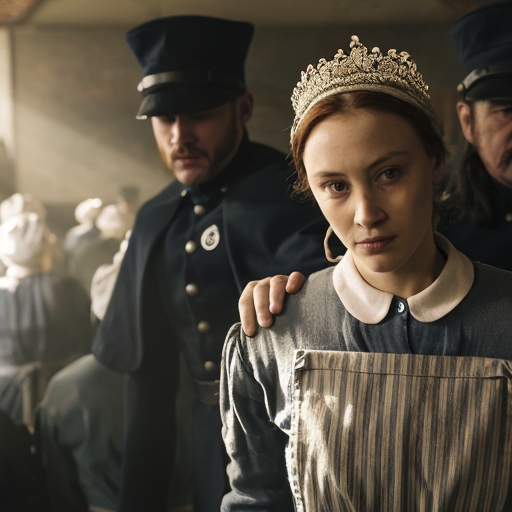}} \\
            & & & \multicolumn{2}{c}{\tiny{``Crowned alias Grace. (Photo by [...]/Netflix)"}} & & & \multicolumn{2}{c}{\tiny{``add a crown"}}\\
        \includegraphics[width=.18\linewidth]{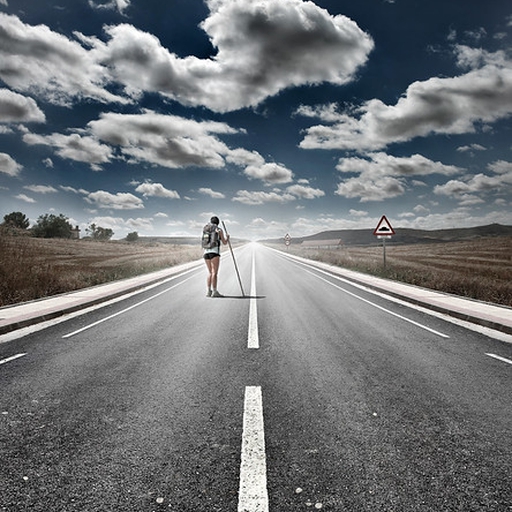} & & &%
        \multicolumn{2}{c}{\includegraphics[width=.18\linewidth]{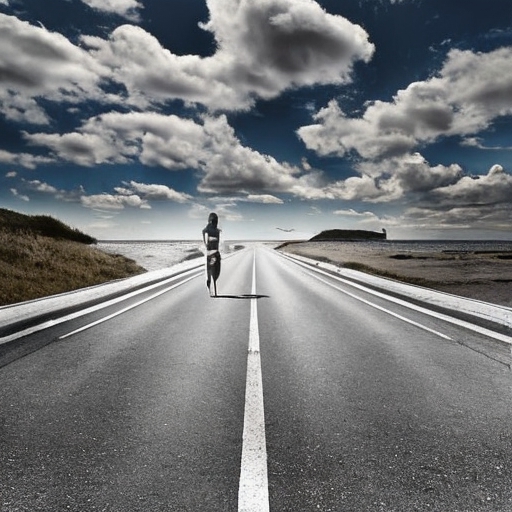} %
        \includegraphics[width=.18\linewidth]{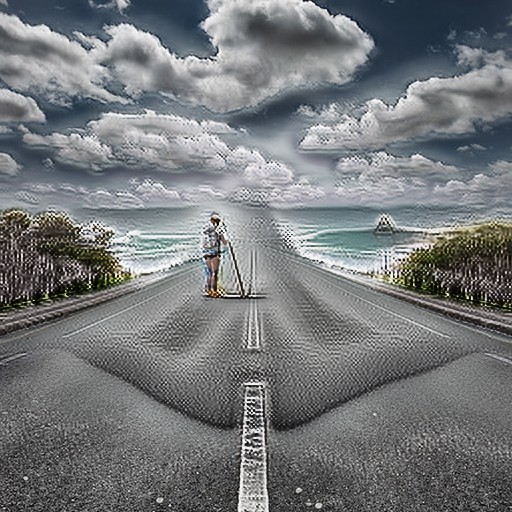}} & & &%
        \multicolumn{2}{c}{\includegraphics[width=.18\linewidth]{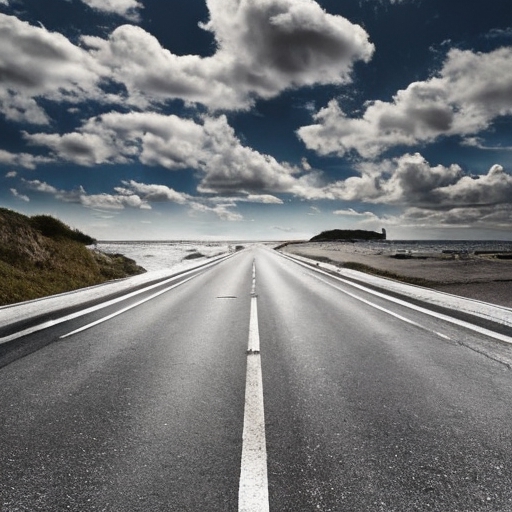} %
        \includegraphics[width=.18\linewidth]{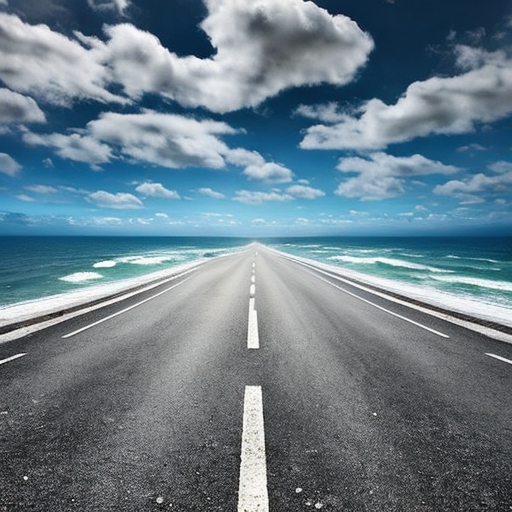}} \\
                   & & & \multicolumn{2}{c}{\tiny{``The Road Leads to the Ocean by Ben Heine"}} & & & \multicolumn{2}{c}{\tiny{``have the road lead to the ocean"}}\\ 
        \includegraphics[width=.18\linewidth]{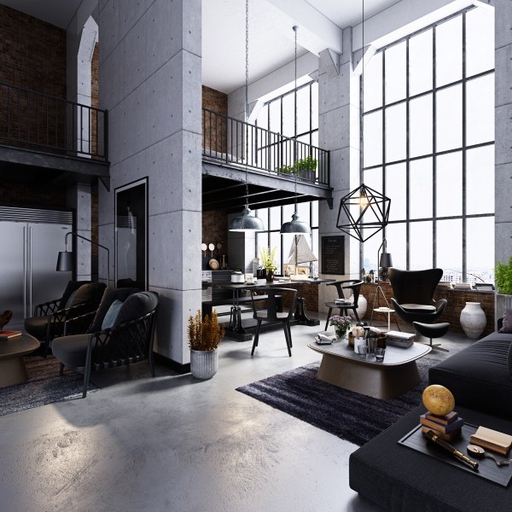} & & &%
        \multicolumn{2}{c}{\includegraphics[width=.18\linewidth]{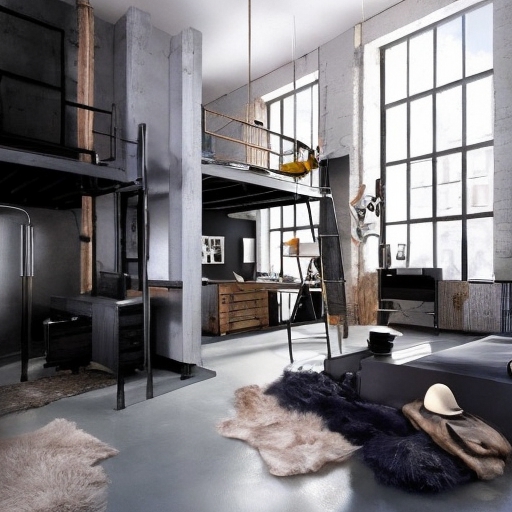} %
        \includegraphics[width=.18\linewidth]{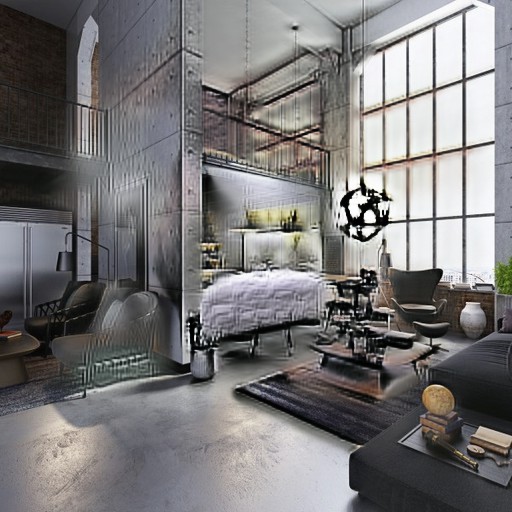}} & & &%
        \multicolumn{2}{c}{\includegraphics[width=.18\linewidth]{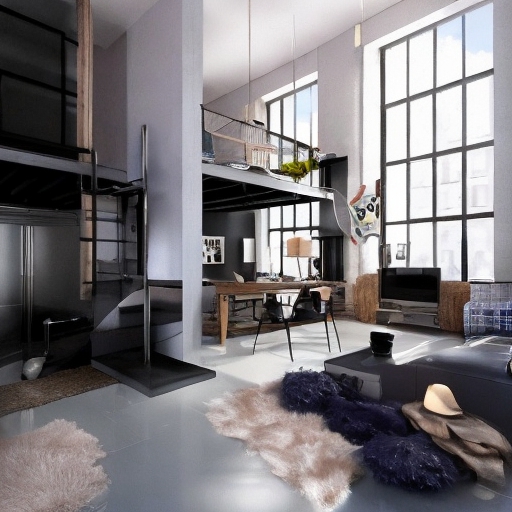} %
        \includegraphics[width=.18\linewidth]{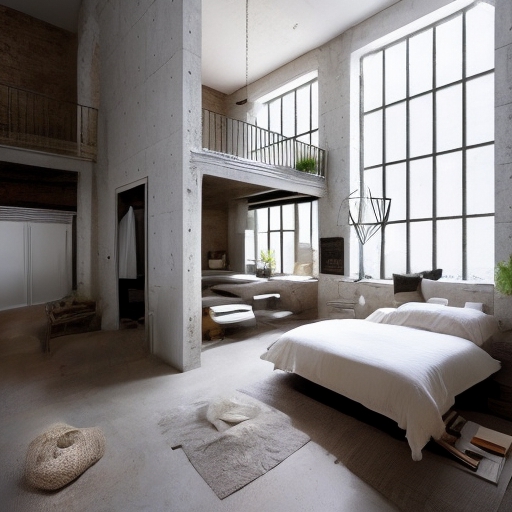}} \\
                    & & & \multicolumn{2}{c}{\tiny{``Industrial design bedroom furniture..."}} & & & \multicolumn{2}{c}{\tiny{``add a bedroom"}}\\
    \end{tabular}
    \vspace{-.1in}
    \caption{Comparison with other editing methods. The input is transformed either by edit string (last two columns) or the ground-truth output image caption (middle two columns). We compare our method against two recent works, SDEdit~\cite{meng2021sdedit} and Text2Live~\cite{bar2022text2live}. We show SDEdit in two configurations: conditioned on the output caption (OP) and conditioned on the edit string (E).}
    \label{fig:comparison}
    \vspace{-2mm}
\end{figure}

\subsection{Baseline comparisons}
\label{subsec:baselines}
We provide qualitative comparisons with SDEdit~\cite{meng2021sdedit} and Text2Live~\cite{bar2022text2live}, as well as quantitative comparisons with SDEdit.
SDEdit~\cite{meng2021sdedit} is a technique for editing images with a pretrained diffusion model, where a partially noised image is passed as input and denoised to produce a new edited image. We compare with the public Stable Diffusion implementation of SDEdit. 
Text2Live~\cite{bar2022text2live} is a technique for editing images by generating a color+opacity augmentation layer, conditioned on a text prompt. We compare with the public implementation released by the authors.

We compare with both methods qualitatively in Figure~\ref{fig:comparison}. We notice that while SDEdit works reasonably well for cases where content remains approximately constant and style is changed, it struggles to preserve identity and isolate individual objects, especially when larger changes are desired. Additionally, it requires a full output description of the desired image, rather than an editing instruction. On the other hand, while Text2Live is able to produce convincing results for edits involving additive layers, its formulation limits the categories of edits that it can handle. 

Quantitative comparisons with SDEdit are shown in Figure~\ref{fig:baselines}. We plot the tradeoff between two metrics, cosine similarity of CLIP image embeddings (how much the edited image agrees with the input image) and the directional CLIP similarity introduced by~\cite{gal2022stylegan} (how much the change in text captions agrees with the change in the images). These are competing metrics---increasing the degree to which the output images correspond to a desired edit will reduce their similarity (consistency) with the input image. Still, we find that when comparing our method with SDEdit, our results have notably higher image consistency (CLIP image similarity) for the same directional similarity values. 

\begin{figure}[t]
    \centering
    \includegraphics[width=\linewidth]{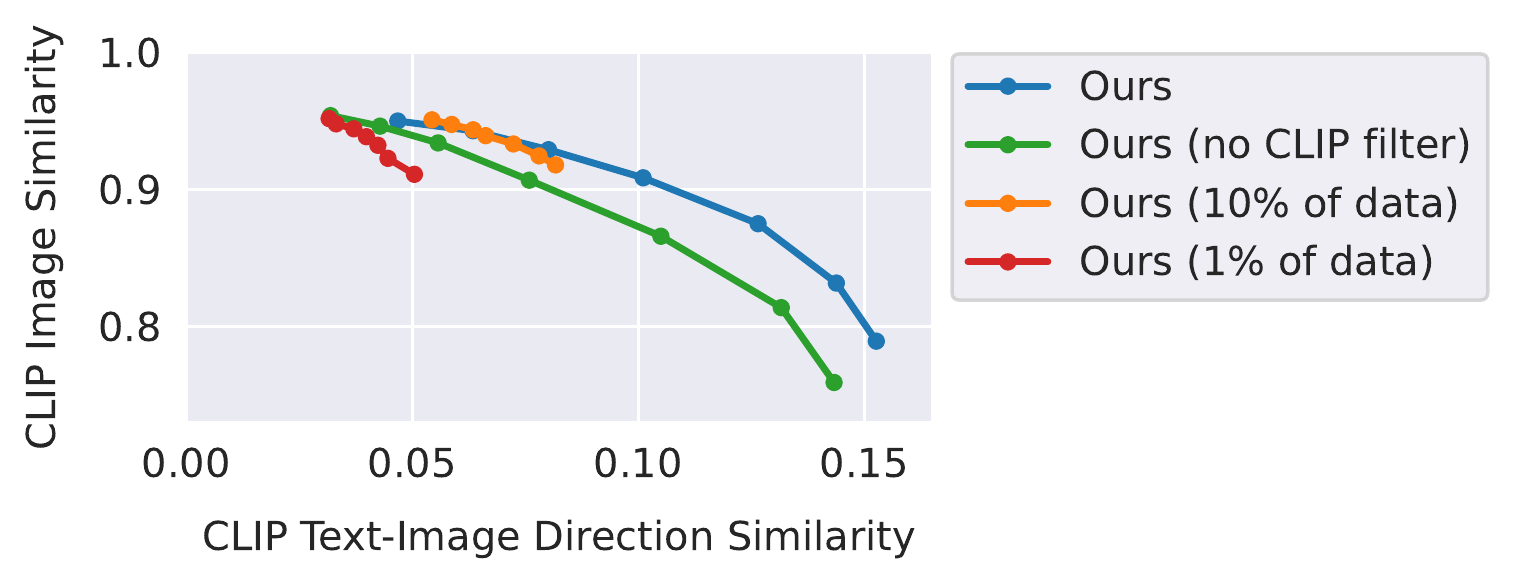}%
        \vspace{-.1in}
    \caption{We compare ablated variants of our model (smaller training dataset, no CLIP filtering) by fixing $s_T$ and sweeping values of $s_I\in[1.0,2.2]$. Our proposed configuration performs best. }
    \label{fig:ablations}
    \vspace{-2mm}
\end{figure}

\begin{figure*}[h!]
    \begin{subfigure}[t]{0.205\linewidth}
        \includegraphics[trim={45mm 0 45mm 0},clip,width=0.491\linewidth]{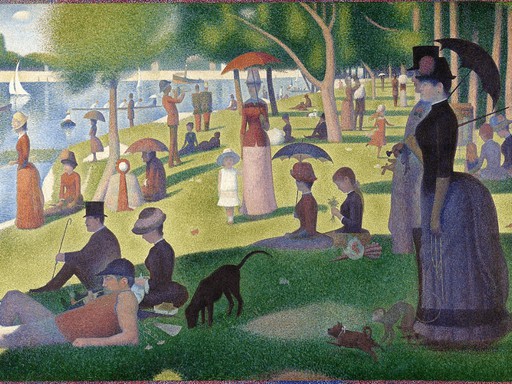}%
        \hfill%
        \includegraphics[trim={45mm 0 45mm 0},clip,width=0.491\linewidth]{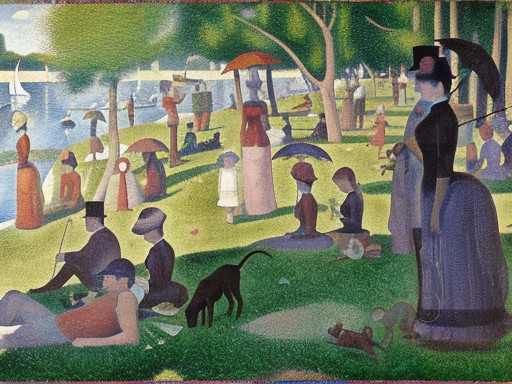}%
        \begin{center}
        \vspace{-2.5mm}
        \scriptsize \emph{``Zoom into the image"}
        \vspace{-1.5mm}
        \end{center}
    \end{subfigure}%
    \hfill%
        \begin{subfigure}[t]{0.205\linewidth}
        \includegraphics[trim={30mm 0 30mm 0},clip,width=0.491\linewidth]{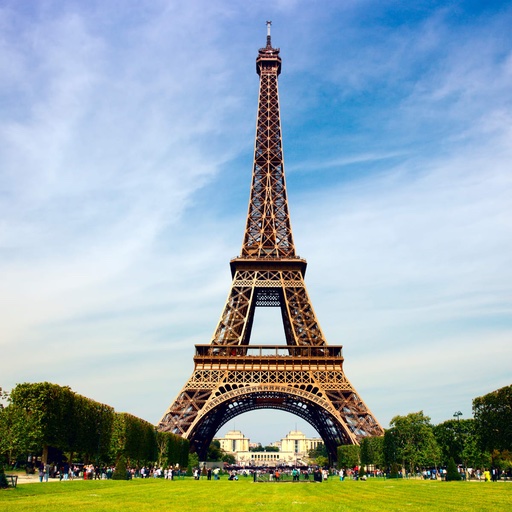}%
        \hfill%
        \includegraphics[trim={30mm 0 30mm 0},clip,width=0.491\linewidth]{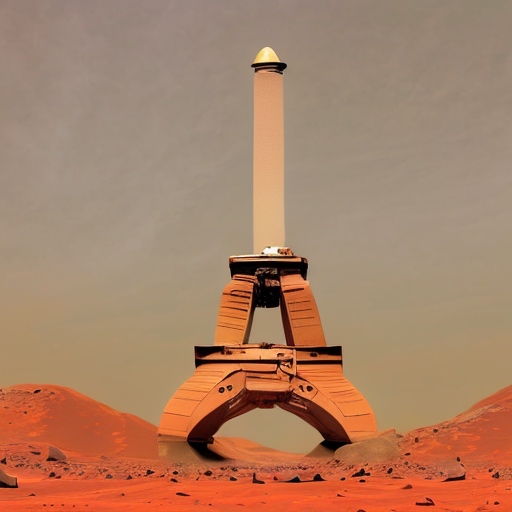}%
        \begin{center}
        \vspace{-2.5mm}
        \scriptsize \emph{``Move it to Mars"}
        \vspace{-1.5mm}
        \end{center}
    \end{subfigure}%
    \hfill%
    \begin{subfigure}[t]{0.238\linewidth}
        \includegraphics[width=0.492\linewidth]{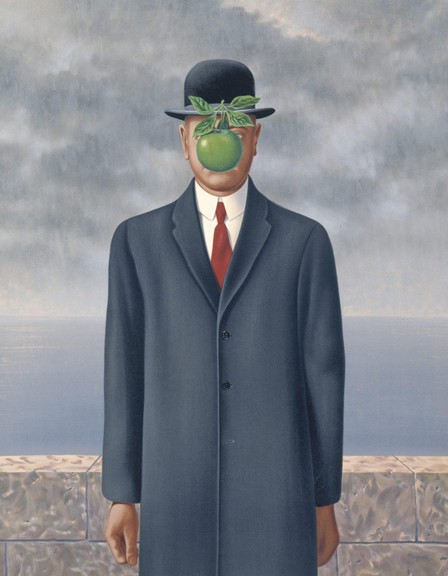}%
        \hfill%
        \includegraphics[width=0.492\linewidth]{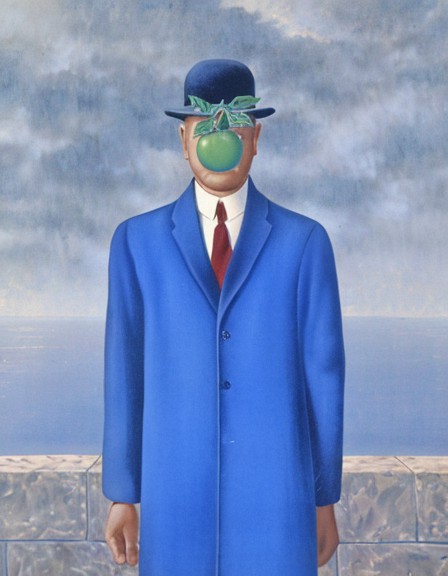}%
        \begin{center}
        \vspace{-2.5mm}
        \scriptsize \emph{``Color the tie blue"}
        \vspace{-1.5mm}
        \end{center}
    \end{subfigure}%
    \hfill%
    \begin{subfigure}[t]{0.305\linewidth}
        \includegraphics[width=0.4935\linewidth]{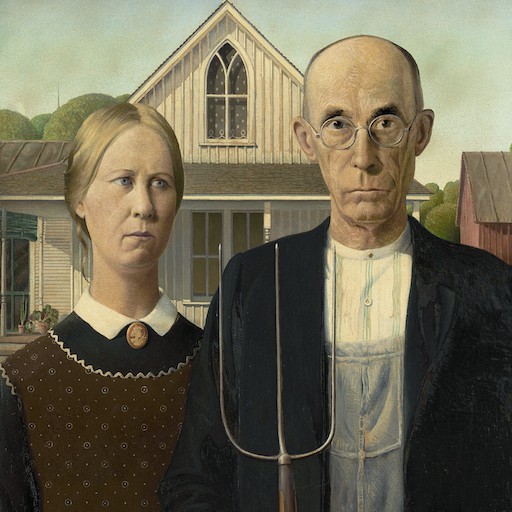}%
        \hfill%
        \includegraphics[width=0.4935\linewidth]{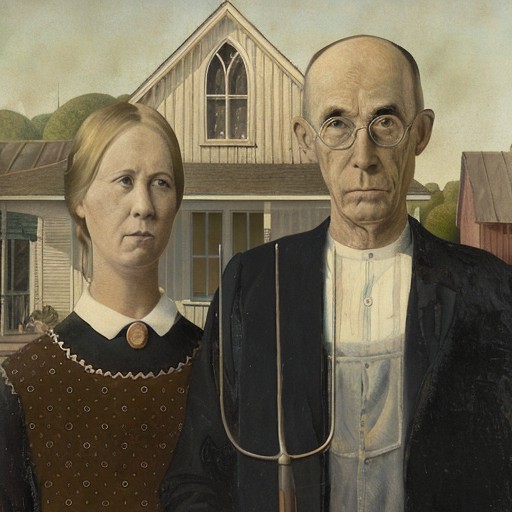}%
        \begin{center}
        \vspace{-2.5mm}
        \scriptsize \emph{``Have the people swap places"}
        \vspace{-1.5mm}
        \end{center}
    \end{subfigure}%
    \vspace{-.5mm}
    \caption{Failure cases. Left to right: our model is not capable of performing viewpoint changes, can make undesired excessive changes to the image, can sometimes fail to isolate the specified object, and has difficulty reorganizing or swapping objects with each other.}
    \vspace{1mm}
    \label{fig:failure_cases}
\end{figure*}

\vspace{-0.5mm}
\subsection{Ablations}
\label{sec:ablations}
In Figure~\ref{fig:ablations}, we provide quantitative ablations for both our choice of dataset size and our dataset filtering approach described in Section~\ref{subsec:data}. We find that decreasing the size of the dataset typically results in decreased ability to perform larger (i.e., more significant) image edits, instead only performing subtle or stylistic image adjustments (and thus, maintaining a high image similarity score, but a low directional score). Removing the CLIP filtering from our dataset generation has a different effect: 
the overall image consistency with the input image is reduced.

We also provide an analysis of the effect of our two classifier-free guidance scales in Figure~\ref{fig:cfg}. Increasing $s_T$ results in a stronger edit applied to the image (i.e., the output agrees more with the instruction), and increasing $s_I$ can help preserve the spatial structure of the input image (i.e., the output agrees more with the input image). We find that values of $s_T$ in the range $5\tight{-}10$ and values of $s_I$ in the range $1\tight{-}1.5$ typically produce the best results. In practice, and for the results shown in the paper, we find it beneficial to adjust guidance weights for each example to get the best balance between consistency and edit strength. 

\section{Discussion}
\label{sec:conclusions}

We demonstrate an approach that combines two large pretrained models, a large language model and a text-to-image model, to generate a dataset for training a diffusion model to follow written image editing instructions.
While our method is able to produce a wide variety of compelling edits to images, including style, medium, and other contextual changes, there still remain a number of limitations. 

\begin{figure*}[h!]
    \centering
    \begin{subfigure}[t]{0.327\linewidth}
        \includegraphics[width=\linewidth]{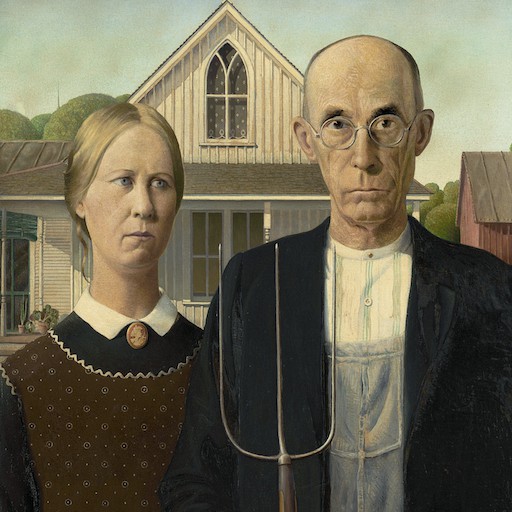}%
        \begin{center}
        \vspace{-2.5mm}
        \scriptsize Input
        \vspace{-1.5mm}
        \end{center}
    \end{subfigure}%
    \hfill%
    \begin{subfigure}[t]{0.327\linewidth}
        \includegraphics[width=\linewidth]{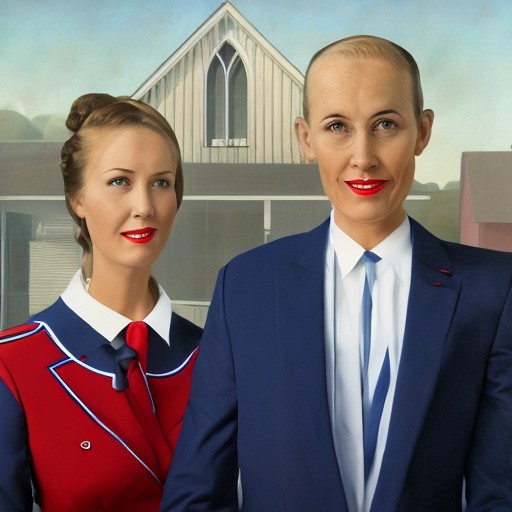}%
        \begin{center}
        \vspace{-2.5mm}
        \scriptsize \emph{``Make them look like flight attendants"}
        \vspace{-1.5mm}
        \end{center}
    \end{subfigure}%
    \hfill%
    \begin{subfigure}[t]{0.327\linewidth}
        \includegraphics[width=\linewidth]{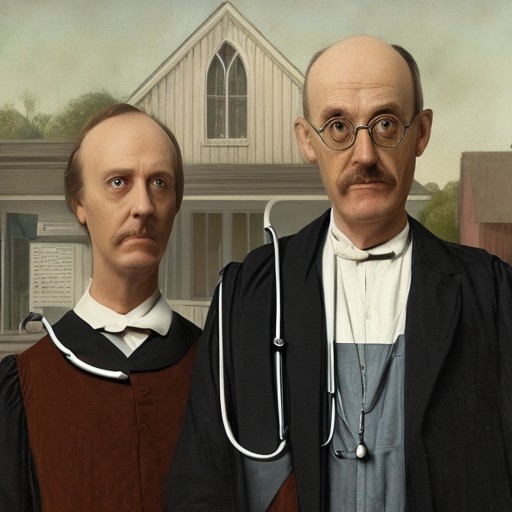}%
        \begin{center}
        \vspace{-2.5mm}
        \scriptsize \emph{``Make them look like doctors"}
        \vspace{-1.5mm}
        \end{center}
    \end{subfigure}%
    \caption{Our method reflects biases from the data and models it is based upon, such as correlations between profession and gender.}
    \vspace{1mm}
    \label{fig:biases}
\end{figure*}

Our model is limited by the visual quality of the generated dataset, and therefore by the diffusion model used to generate the imagery (in this case, Stable Diffusion~\cite{rombach2022high}). Furthermore, our method's ability to generalize to new edits and make correct associations between visual changes and text instructions is limited by the human-written instructions used to fine-tune GPT-3~\cite{brown2020language}, by the ability of GPT-3 to create instructions and modify captions, and by the ability of Prompt-to-Prompt~\cite{hertz2022prompt} to modify generated images. In particular, our model struggles with counting numbers of objects and with spatial reasoning (e.g., \emph{``move it to the left of the image''}, \emph{``swap their positions''}, or \emph{``put two cups on the table and one on the chair''}), just as in Stable Diffusion and Prompt-to-Prompt. Examples of failures can be found in Figure~\ref{fig:failure_cases}.
Furthermore, there are well-documented biases in the data and the pretrained models that our method is based upon, and therefore the edited images from our method may inherit these biases or introduce other biases (Figure~\ref{fig:biases}). 

Aside from mitigating the above limitations, our work also opens up questions, such as: how to follow instructions for spatial reasoning, how to combine instructions with other conditioning modalities like user interaction, and how to evaluate instruction-based image editing. Incorporating human feedback to improve the model is another important area of future work, and strategies like human-in-the-loop reinforcement learning could be applied to improve alignment between our model and human intentions.

\paragraph{Acknowledgments} We thank Ilija Radosavovic, William Peebles, Allan Jabri, Dave Epstein, Kfir Aberman, Amanda Buster, and David Salesin. Tim Brooks is funded by an NSF Graduate Research Fellowship. Additional funding provided by 
a research grant from SAP and a gift from Google. 

{\small
\bibliographystyle{ieee_fullname}
\bibliography{PaperForReview}
}

\clearpage
\appendix
\section{Implementation Details}
\label{subsec:implementation}

\subsection{Instruction and Caption Generation}
We finetune GPT3 to generate edit instructions and edited captions. The text prompt used during fine-tuning is the input caption concatenated with \texttt{"\textbackslash n\#\#\textbackslash n"} as a separator token. The text completion is a concatenation of the instruction and edited caption with \texttt{"\textbackslash n\%\%\textbackslash n"} as a separator token in between the two and \texttt{"\textbackslash nEND"} appended to the end as the stop token. During inference, we sample text completions given new input captions using \texttt{temperature=0.7} and \texttt{frequency\_penalty=0.1}. We exclude generations where the input and output captions are the same.

\subsection{Paired Image Generation}
We generate paired before/after training images from paired before/after captions using Stable Diffusion~\cite{rombach2022high} in combination with Prompt-to-Prompt~\cite{hertz2022prompt}. We use exponential moving average (EMA) weights of the Stable Diffusion v1.5 checkpoint and the improved ft-MSE autoencoder weights. We generate images with 100 denoising steps using an Euler ancestral sampler with denoising variance schedule proposed by Kerras \emph{et al.}~\cite{karras2022elucidating}. We ensure the same latent noise is used for both images in each generated pair (for initial noise as well as noise introduced during stochastic sampling).

Prompt-to-Prompt replaces cross-attention weights in the second generated image differently based on the specific edit type: word swap, adding a phrase, increasing or decreasing weight of a word. We instead replaced \emph{self}-attention weights of the second image for the first $p$ fraction of steps, and use the same attention weight replacement strategy for all edits.

We generation $100$ pairs of images for each pair of captions. We filter training data for an image-image CLIP threshold of 0.75 to ensure images are not too different, an image-caption CLIP threshold of 0.2 to ensure images correspond with their captions, and a directional CLIP similarity of 0.2 to ensure the change in before/after captions correspond with the change in before/after images. For each each pair of captions, we sort any image pairs that pass all filters by the directional CLIP similarity and keep up to 4 examples.

\subsection{Training InstructPix2Pix}
We train our image editing model for 10,000 steps on $8\times$ 40GB NVIDIA A100 GPUs over $25.5$ hours. We train at $256\times256$ resolution with a total batch size of 1024. We apply random horizontal flip augmentation and crop augmentation where images are first resized randomly between 256 and 288 pixels and then cropped to 256. We use a learning rate of $10^{-4}$ (without any learning rate warm up). We initialize our model from EMA weights of the Stable Diffusion v1.5 checkpoint, and adopt other training settings from the public Stable Diffusion code base. 

While our model is trained at $256\times256$ resolution, we find it generalized well to $512\times512$ resolution at inference time, and generate results in this paper at $512$ resolution with 100 denoising steps using an Euler ancestral sampler with denoising variance schedule proposed by Kerras \emph{et al.}~\cite{karras2022elucidating}. Editing an image with our model takes roughly $9$
seconds on an A100 GPU.


\section{Classifier-free Guidance Details}
\label{sec:cfg2a}

As discussed in Section~\ref{subsubsec:cfg}, we apply classifier-free guidance with respect to two conditionings: the input image $c_I$ and the text instruction $c_T$. We introduce separate guidance scales $s_I$ and $s_T$ that enable separately trading off the strength of each conditioning. Below is the modified score estimate for our model with classifier-free guidance (copied from Equation~\ref{eq:cfg2}):

\vspace{-2.5mm}
\begin{equation*}
\begin{split}
    \tilde{e_{\theta}}(z_t, c_I, c_T) = &\: e_{\theta}(z_t, \varnothing, \varnothing) \\ &+ s_I \cdot (e_{\theta}(z_t, c_I, \varnothing) - e_{\theta}(z_t, \varnothing, \varnothing)) \\ &+ s_T \cdot (e_{\theta}(z_t, c_I, c_T) - e_{\theta}(z_t, c_I, \varnothing))
\end{split}
\end{equation*}
\vspace{0.1mm}

Our generative model learns $P(z|c_I,c_T)$, the probability distribution of image latents $z = \mathcal{E}(x)$ conditioned on an input image $c_I$ and a text instruction $c_T$. We arrive at our particular classifier-free guidance formulation by expressing the conditional probability as follows:

\vspace{-2.5mm}
\begin{equation*}
    P(z|c_T,c_I) = \frac{P(z,c_T,c_I)}{P(c_T,c_I)} = \frac{P(c_T|c_I,z)P(c_I|z)P(z)}{P(c_T,c_I)}
\end{equation*}
\vspace{0.1mm}

Diffusion models estimate the score~\cite{hyvarinen2005estimation} of the data distribution, i.e., the derivative of the log probability. Taking the logarithm gives us the following expression:

\vspace{-2.5mm}
\begin{equation*}
\begin{split}
\log(P(z|c_T,c_I)) = &\: \log(P(c_T|c_I,z)) + \log(P(c_I|z)) \\ &+ \log(P(z)) - \log(P(c_T,c_I))
\end{split}
\end{equation*}
\vspace{0.1mm}

\noindent Taking the derivative and rearranging we attain:

\vspace{-2.5mm}
\begin{equation*}
\begin{split}
\nabla_z\log(P(z|c_T,c_I)) = &\: \nabla_z\log(P(z)) \\ &+ \nabla_z\log(P(c_I|z)) \\ &+ \nabla_z\log(P(c_T|c_I,z))
\end{split}
\end{equation*}
\vspace{0.1mm}

This corresponds with the terms in our classifier-free guidance formulation in Equation~\ref{eq:cfg2}. Our guidance scale $s_I$ effectively shifts probability mass toward data where an implicit classifier $p_{\theta}(c_I|z_t)$ assigns high likelihood to the image conditioning $c_I$, and our guidance scale $s_T$ effectively shifts probability mass toward data where an implicit classifier $p_{\theta}(c_T|c_I,z_t)$ assigns high likelihood to the text instruction conditioning $c_T$. Our model is capable of learning these implicit classifiers by taking the differences between estimates with and without the respective conditional input. Note there are multiple possible formulations such as switching the positions of $c_T$ and $c_I$ variables. We found that our particular decomposition works better for our use case in practice.

\end{document}